\documentclass[sn-basic,iicol]{sn-jnl}

\usepackage{graphicx}%
\usepackage{multirow}%
\usepackage{amsmath,amssymb,amsfonts}%
\usepackage{amsthm}%
\usepackage{mathrsfs}%
\usepackage{bbm}
\usepackage[title]{appendix}%
\usepackage{xcolor}%
\usepackage{textcomp}%
\usepackage{manyfoot}%
\usepackage{booktabs}%
\usepackage{algorithm}%
\usepackage{algorithmicx}%
\usepackage{algpseudocode}%
\usepackage{listings}%
\usepackage{natbib}
\usepackage{fancyhdr}
\usepackage{multicol}
\usepackage{pdflscape}
\usepackage{tabularx}

\usepackage{tikz}
\usetikzlibrary{calc}
\usepackage{atbegshi} 

\AtBeginShipout{
\AtBeginShipoutAddToBox{
    \begin{tikzpicture}[remember picture, overlay]
    \node[draw] at ($(current page.north)-(0,0.75)$) {APPROVED FOR PUBLIC RELEASE};
    \end{tikzpicture}
    \begin{tikzpicture}[remember picture, overlay]
    \node[draw] at ($(current page.south)+(0,0.75)$) {APPROVED FOR PUBLIC RELEASE};
    \end{tikzpicture}
}
}

\begin{document}

\title[Article Title]{Are Deep Learning Models Robust to Partial Object Occlusion in Visual Recognition Tasks?}


\author*[1]{\fnm{Kaleb} \sur{Kassaw}}\email{kaleb.kassaw@duke.edu}

\author[1]{\fnm{Francesco} \sur{Luzi}}\email{francesco.luzi@duke.edu}

\author[1]{\fnm{Leslie M.} \sur{Collins}}\email{leslie.collins@duke.edu}

\author[2]{\fnm{Jordan M.} \sur{Malof}}\email{jmdrp@missouri.edu}

\affil[1]{\orgdiv{Department of Electrical and Computer Engineering}, \orgname{Duke University}, \orgaddress{\city{Durham}, \state{North Carolina}, \postcode{27708}, \country{United States}}}

\affil[2]{\orgdiv{Department of Electrical Engineering and Computer Science}, \orgname{University of Missouri}, \orgaddress{\city{Columbia}, \state{Missouri}, \postcode{65211}, \country{United States}}}

\abstract{Image classification models, including convolutional neural networks (CNNs), perform well on a variety of classification tasks but struggle under conditions of partial occlusion, i.e., conditions in which objects are partially covered from the view of a camera. Methods to improve performance under occlusion, including data augmentation, part-based clustering, and more inherently robust architectures, including Vision Transformer (ViT) models, have, to some extent, been evaluated on their ability to classify objects under partial occlusion. However, evaluations of these methods have largely relied on images containing artificial occlusion, which are typically computer-generated and therefore inexpensive to label. Additionally, methods are rarely compared against each other, and many methods are compared against early, now outdated, deep learning models. We contribute the Image Recognition Under Occlusion (IRUO) dataset, based on the recently developed Occluded Video Instance Segmentation (OVIS) dataset (\cite{qi_occluded_2022}). IRUO utilizes real-world and artificially occluded images to test and benchmark leading methods’ robustness to partial occlusion in visual recognition tasks. In addition, we contribute the design and results of a human study using images from IRUO that evaluates human classification performance at multiple levels and types of occlusion, including diffuse occlusion. We find that modern CNN-based models show improved recognition accuracy on occluded images compared to earlier CNN-based models, and ViT-based models are more accurate than CNN-based models on occluded images, performing only modestly worse than human accuracy. We also find that certain types of occlusion, including diffuse occlusion, where relevant objects are seen through "holes" in occluders such as fences and leaves, can greatly reduce the accuracy of deep recognition models as compared to humans, especially those with CNN backbones.} 


\maketitle

\section{Introduction}\label{sec:intro}

\begin{figure}[t]
\begin{center}
\begin{tabular}{ccc}
\includegraphics[width=0.3\linewidth]{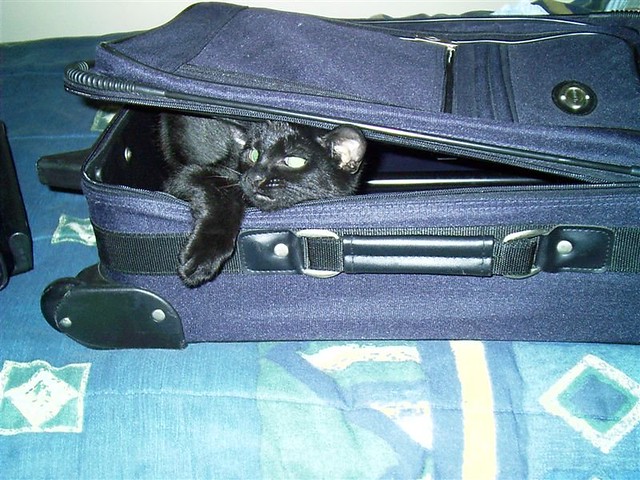}&
\includegraphics[width=0.3\linewidth]{333772.jpg}&
\includegraphics[width=0.3\linewidth]{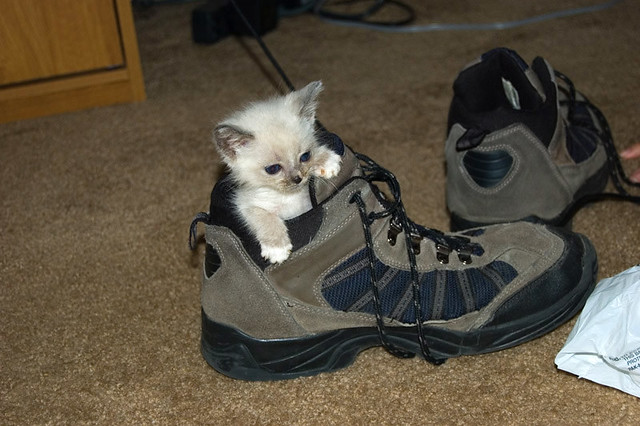}\\
(a)&(b)&(c)
\end{tabular}

\caption{Images from Microsoft's COCO dataset (\cite{lin_microsoft_2015}), a popular dataset for visual recognition tasks including object detection, demonstrating partial object occlusion. In each of these images, the object class "cat" is partially occluded by (a) a suitcase, among the object classes listed in COCO, (b) another cat, and (c) a boot, a class of object not among the object clases listed in COCO}
\label{fig:intro_occlusion}
\end{center}
\end{figure}

Deep learning models, particularly deep neural networks (DNNs), have demonstrated tremendous success on visual recognition tasks, even matching or exceeding human performance on some benchmarks (e.g., \cite{xie_aggregated_2017,dosovitskiy_image_2021,liu_swin_2021}). However, one important real-world condition in which DNNs still struggle is occlusion, wherein the target object is partially obscured, or occluded, by other objects in the scene. Fig. \ref{fig:intro_occlusion} illustrates a scenario where a target object, of the class \textit{cat}, is occluded to varying degrees by scene content, including other objects of the target class. In this work we aim to address two important and fundamental open questions regarding image recognition under occlusion. 

\textbf{(i) Which existing models are most accurate on occluded imagery?}

The answer to this question is useful for practitioners facing occlusion in imagery and would provide guidance for future research. Several recent publications have proposed occlusion-robust recognition models (see Sec. \ref{sec:related}), and demonstrated that their proposed approaches improve visual recognition accuracy over conventional DNNs. However, these comparisons, e.g., in \cite{kortylewski_combining_2020, xiao_tdapnet_2019, cen_deep_2021}, suffer from several limitations, e.g., comparisons to only early deep learning models such as AlexNet (\cite{krizhevsky_imagenet_2012}) and VGG (\cite{simonyan_very_2015}) and other models expressly designed for occluded scenes, datasets containing few classes (e.g., vehicles), and datasets with few examples of real-world occlusion. This is an especially important set of comparisons to make, as deep learning has advanced significantly since these early models, greatly improving top-1 classification accuracy on large-scale datasets such as ImageNet (\cite{deng_imagenet_2009}). Therefore, to our knowledge, no existing work has performed a rigorous empirical comparison -- or benchmark -- of both existing conventional DNNs and/or recent occlusion-robust models. 

\textbf{(ii) Are existing models robust to occlusion? Human and model comparison}

For a model to be "robust" to occlusion, we mean that it does not degrade much faster than necessary (e.g., compared to an optimal model) in the presence of occlusion. Numerous publications have demonstrated that DNN recognition accuracy degrades in the presence of occlusion (\cite{zhu_robustness_2019,kortylewski_compositional_2020,cen_deep_2021,kortylewski_combining_2020,xiao_tdapnet_2019}). Note however that the accuracy of any recognition model -- even an optimal model -- might \textit{necessarily} degrade in the presence of occlusion, due to loss of image information. Therefore it is unclear whether or not the accuracy degradation observed in existing research can be mitigated through improved modeling and, therefore, whether existing models are robust to occlusion. 

To evaluate whether a particular model is occlusion-robust, one must compare against an optimal model, yet estimating such a model's performance is difficult. Human visual recognition accuracy is often used as such an estimator, and one recent publication has compared human and DNN-based recognition accuracy as occlusion increases (\cite{zhu_robustness_2019}). This study concluded that DNN-based models do indeed degrade more quickly than human observers, and it provides one occlusion-robustness estimate (i.e., how much performance do we stand to gain with research investment). Although this work provides valuable insights, the test dataset used for this study (based on the \textit{VehicleOcclusion} dataset in \cite{wang_detecting_2017}) suffers from several limitations that undermine the ability to draw generalizable conclusions. First, the testing dataset is limited in both size (e.g., only 500 test images) and variability (e.g., only vehicles classes). Second, the authors predominantly use synthetic occlusions consisting of highly unrealistic objects and scenarios, and it is unclear if such scenarios are a good proxy for real-world occlusions. Lastly, the work utilized relatively old models (e.g., AlexNet (\cite{krizhevsky_imagenet_2012}), VGG (\cite{simonyan_very_2015}) that exhibit inferior performance to modern models, and as we confirm in this work, also yield inferior performance under occlusion. 

\subsection{Contributions of This Work}
In this work we attempt to provide comprehensive answers to the aforementioned questions. First, we curated a large and diverse dataset of real-world occlusions termed the Image Recognition Under Occlusion (IRUO) dataset, which we built upon the Occluded Video Instance Segmentation (OVIS) dataset (\cite{qi_occluded_2022}). The IRUO contains 23 classes of objects and 88 thousand images and addresses several limitations associated with previous datasets, such as their limited size, limited target object diversity, or their lack of real-world occlusion. Using IRUO, we address question (i) by comparing state-of-the-art representatives from each of three relevant model types: convolutional models (e.g., ResNeXt (\cite{xie_aggregated_2017})), Vision Transformers (e.g., Swin (\cite{liu_swin_2021})), and models that are especially designed to be occlusion robust (e.g., CompositionalNet (\cite{kortylewski_combining_2020})). We also leverage our benchmark results to examine whether the usage of synthetically occluded imagery (e.g., in Fig. \ref{fig:our_artificial_occlusions}) is an accurate proxy for real-world occlusion. Synthetic occlusions are much easier to generate and control, and therefore a positive answer to this question would empower future investigation of occlusion, as well as help validate historical studies that made use of synthetic occlusion. To address question (ii) we obtain classification predictions from twenty human observers on our IRUO dataset, and compare their accuracy under occlusion against all of the data-driven models in our study.

Contributions of this work can be summarized as follows. 
\begin{enumerate}
    \item The Image Recognition Under Occlusion (IRUO) Dataset: the first large-scale public benchmark for image recognition under occlusion). This dataset is built upon the OVIS dataset (\cite{qi_occluded_2022}) and comprises unprecedented size and diversity. 
    \item A rigorous comparison of state-of-the-art image classification models under varying levels of occlusion. This includes convolutional models, Vision Transformers, and models specifically designed for occlusion.
    \item A rigorous comparison of model and human recognition accuracy on our benchmark, making it possible to determine whether, and to what extent, existing DNN-based models can be made more robust to occlusion.
    \item An investigation of whether synthetically occluded imagery (e.g., see Fig. \ref{fig:our_artificial_occlusions}) is an accurate  proxy for real-world occlusion. 
\end{enumerate}

To support future research, we publish all of our datasets, models, and results.\footnote{https://github.com/kalebkassaw/iruo}

\section{Related Work}
\label{sec:related}

\textbf{Existing occlusion-robust methods.} Recent methods have been proposed to address the decrease in model accuracy under conditions of occlusion. These methods include data augmentation (\cite{cen_deep_2021,yun_cutmix_2019,chen_transmix_2022}), part-based modeling techniques (\cite{kortylewski_combining_2020,kortylewski_compositional_2020,xiao_tdapnet_2019,xiao_tdmpnet_2022,yang_robust_2018}), and development of architectures inherently more robust to partial occlusion (\cite{dosovitskiy_image_2021,naseer_intriguing_2021}). Several methods of data augmentation are used to encourage models to focus on image parts. Mixup and CutMix are augmentations that replace single training images with combinations of multiple images, through weighted averages over all pixels (Mixup) or cut-and-paste image areas (CutMix). Cutout (\cite{devries_improved_2017}), a closely related augmentation to CutMix, does not add pieces of a second image; instead, cutouts are replaced with gray boxes. The authors in each paper hypothesize that, by forcing deep networks to make inferences on parts, they can confer part-based knowledge to these models. In \cite{chen_transmix_2022}, in proposing TransMix, an extension of CutMix that re-weights proportions of images based on relative transformer attention values, the authors hypothesize that this part-based knowledge is linked to improved performance under occlusion. Data augmentation techniques, including CutMix 
and TransMix, are hypothesized to improve recognition accuracy on occluded images as they splice parts of images and encourage models to learn labels that are proportional to image area. Data augmentation has also been proposed at the convolutional deep feature level to improve recognition accuracy under occlusion (\cite{cen_deep_2021}); this method collects differences in deep vectors between unoccluded and occluded objects to confer occlusion robustness to models. Part-based modeling techniques have been used to improve performance under some occluded scenes while preserving the convolutional backbones of many deep learning models; CompositionalNet (\cite{kortylewski_compositional_2020}), TDAPNet (\cite{xiao_tdapnet_2019}), and TDMPNet (\cite{xiao_tdmpnet_2022}) cluster deep feature representations in high-dimensional space and use various methods to assign these features to either object classes or occluders, explicitly for the purpose of robustness to occlusion. Although not explicitly designed for the purpose, the Vision Transformer (\cite{dosovitskiy_image_2021}) has been shown to perform well under conditions of constant-valued occlusions coinciding with input image patches (\cite{naseer_intriguing_2021}). We include CompositionalNet in our experiments, in addition to Deep Feature Augmentation, as these methods have been shown to improve performance under some conditions of occlusion, compared to VGG (\cite{simonyan_very_2015}). We additionally include evaluations of CutMix and Mixup (\cite{zhang_mixup_2018}), as they are widely deployed in modern models (e.g., \cite{dosovitskiy_image_2021, touvron_training_2021, liu_swin_2021}) to improve overall recognition accuracy and suggested to improve accuracy under conditions of occlusion (\cite{chen_transmix_2022}).

\textbf{Benchmarking human and algorithm performance under occlusion.} Human visual recognition performance on occlusion has been evaluated in \cite{zhu_robustness_2019}, where accuracy of humans on a five-class vehicle dataset with artificial occlusion is compared to the accuracy of deep learning models including AlexNet (\cite{krizhevsky_imagenet_2012}) and VGG (\cite{simonyan_very_2015}). This study suggests that human performance is not attained by deep learning models on images with objects that are highly occluded, as human performance on highly occluded objects is significantly higher than that of any model tested. However, this study has several limitations; the dataset contains only five vehicle classes, compares against aging models instead of more modern models of the time, e.g., ResNet (\cite{he_deep_2016}), and it is not publicly available.

Recent research has suggested that Vision Transformer-based models may perform better under conditions such as mask occlusion and adversarial patches, suggesting that newer attention-based deep learning models may demonstrate higher performance than convolutional neural networks under conditions of real-world occlusion (\cite{naseer_intriguing_2021}). We include these algorithms in a benchmark of visual recognition accuracy under occlusion, in addition to comparing these results with human accuracy, for the first time.

\section{Background Methods}

\subsection{Models}
\label{sec:models}
 We evaluate several state-of-the-art vision models from three relevant categories: convolutional models, transformers, and models designed specifically for occlusion robustness. Among occlusion-specific models, we include the CompositionalNet (\cite{kortylewski_compositional_2020}), because it was recently shown to outperform other such models (e.g., TDAPNet (\cite{xiao_tdapnet_2019}), and "two-stage voting" (\cite{tang_recurrent_2018, zhu_robustness_2019})). Among convolutional models we include VGG (\cite{simonyan_very_2015}) for continuity with prior work, as VGG was included in several previous benchmarks (e.g., in \cite{zhu_robustness_2019, kortylewski_compositional_2020, xiao_tdapnet_2019}). We also include state-of-the-art convolutional models such as ResNet (\cite{he_deep_2016}) and ResNeXt (\cite{xie_aggregated_2017}). Among state-of-the-art transformer-based models, we include the Vision Transformer (\cite{dosovitskiy_image_2021}), DeiT (\cite{touvron_training_2021}), and Swin (\cite{liu_swin_2021}). 

\subsection{Augmentations}\label{subsec:augmentations}
We also propose the testing of various augmentation and model regularization methods, including Mixup (\cite{zhang_mixup_2018}), CutMix (\cite{yun_cutmix_2019}), and Deep Feature Augmentation (\cite{cen_deep_2021}). Mixup and CutMix are state-of-the-art augmentation methods that are commonly used in leading deep learning models regardless of the presence of occlusion in the images they classify, and their usage is hypothesized to improve model robustness to partial occlusion by conferring knowledge of object parts to models (\cite{chen_transmix_2022}). Deep Feature Augmentation is explicitly designed to improve model performance in scenes of partial occlusion (\cite{cen_deep_2021}) over an otherwise equivalent CNN-based model.

\subsection{Statistical Tests}
\label{sec:stats}

\textbf{Friedman Test (\cite{friedman_use_1937})}. To determine whether or not models are ranked differently by real versus synthetic occlusions (see Sec. \ref{sec:synth}), we use the Friedman test in \cite{friedman_use_1937}. This test assumes that we have a collection of $k$ objects to rank, and that we have $n$ evaluation metrics, or "judges", for ranking the objects. The objective of the test is to determine whether the judges produce random rankings. Specifically, the null hypothesis of this test is that the rankings of each judge are produced by randomly drawing each of the $k$ objects without replacement, and then assigning their rank based upon the order in which they were drawn. Rejection of the null hypothesis then suggests that the rankings of each judge are non-random so that some objects are ranked consistently above others across the judges. To test this hypothesis, the Friedman test prescribes the calculation of a $Q$ statistic based upon the observed $k$ rankings of each of the $n$ available judges. For a sufficiently large $k$, the $Q$ statistic is drawn from a $\chi^2$ distribution with $k-1$ degrees of freedom, and we reject the null hypothesis if Q is above the 95th percentile of this $\chi^2$ distribution (equivalent to $p < 0.05$).

\textbf{Tukey's Rule (\cite{tukey_exploratory_1977})}. In Sec. \ref{sec:human_study} we estimate human recognition accuracy in the presence of occlusion, and we use Tukey's rule (\cite{tukey_exploratory_1977}) to remove unrepresentative human subjects (i.e., outliers) from consideration. Tukey's rule prescribes that data is drawn from the same underlying distribution with a common mean and variance. By this rule, data points in a set falling 1.5 interquartile ranges outside the interquartile range are classified as outliers, and the removal of these points for calculating summary statistics is justified.

\subsection{Human Labeling Procedure}
\label{sec:human_labeling_procedure}

We employ a method of human labeling known as \textit{Scalable Multi-Label Annotation}, first used in \cite{deng_scalable_2014}. This method of data collection attempts to minimize human error due to factors other than strict identification of classes. For a (typically large) set of classes within a dataset, Scalable Multi-Label Annotation asks a series of informative categorical questions to narrow human responses to a much smaller set of classes within the dataset. These categorical questions can be broad at first, e.g., "Is an animal present?". A sample hierarchy reflecting such an iterative procedure in this study is shown in Fig. \ref{fig:hierarchy}. This iterative collection procedure is designed to remove the need for 1-of-n-class labeling (i.e., asking whether or not each class in a dataset is present for every image) while also addressing humans' limited memory (\cite{deng_scalable_2014}). DNN-based models, by nature of utilizing large computational resources, are capable of remembering characteristics distinctive to each class; humans, by comparison, perform best with few-class, hierarchical prompts (\cite{deng_scalable_2014}).

In addition, this hierarchical approach enables the analysis of whether or not humans (and models) are selecting classes that are relatively close (e.g., labeling a tiger in place of a cat) or relatively distant (e.g., labeling a tiger in place of an airplane), as categorical responses can be used in place of fine labels for certain analyses. This allows us to compare the types of errors made by both humans and models beyond simply comparing top-1 accuracy scores.

\section{The Image Recognition Under Occlusion (IRUO) Dataset}\label{sec:dataset}

\begin{figure}[b]
\begin{center}
\begin{tabular}{ccc}
\includegraphics[width=0.27\linewidth]{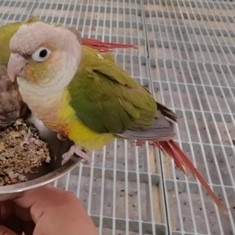}&
\includegraphics[width=0.27\linewidth]{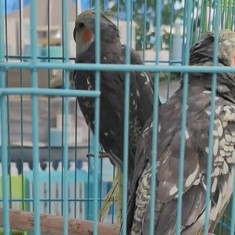}&
\includegraphics[width=0.27\linewidth]{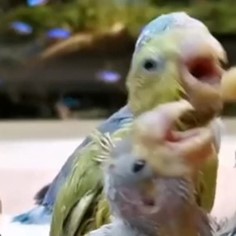}\\
(a)&(b)&(c)
\end{tabular}

\caption{Images showing various levels of real occlusion, increasing from left to right, in the Image Recognition Under Occlusion (IRUO) dataset. We propose to use images including these for evaluation of models’ robustness to occlusion in recognition tasks, owing to realistic occlusions and a wide diversity of scenes. Occlusion levels: (a) 0: no occlusion, (b) 1: some occlusion, in which up to 50 percent of the object is hidden from view, (c) 2: severe occlusion, in which more than 50 percent of an object is hidden from view}
\label{fig:occ_levels}
\end{center}
\end{figure}

We build the IRUO dataset based upon the Occluded Video Instance Segmentation (OVIS) dataset (\cite{qi_occluded_2022}), first used in video segmentation problems. The OVIS dataset is comprised of a training set of approximately 3.5k objects in 600 videos, and validation and test sets of 750 objects in 150 videos each. Each of these objects is labeled with a corresponding occlusion level, (0) no occlusion; (1) some occlusion, in which up to 50 percent of the object is hidden from view; and (2) severe occlusion, in which more than 50 percent of the object is hidden from view. To create IRUO, we needed to curate the OVIS video dataset so that it is appropriate for the task of image classification. We chose to build an image classification dataset (as opposed to segmentation or detection) because, to our knowledge, all occlusion-robust image recognition methods (e.g., CompositionalNet (\cite{kortylewski_compositional_2020}), TDAPNet (\cite{xiao_tdapnet_2019})) are suitable only for the classification task. This was done by cropping individual target objects from OVIS video frames using the bounding boxes provided with OVIS for each target instance. To account for minor bounding box label errors and preserve object aspect ratios, we then expand the shorter dimension from the center to match the longer dimension, expand all dimensions 20 pixels from the center point, and use the subsequent crops to generate our image classification dataset. Sample images are shown in Fig. \ref{fig:occ_levels}, where a parrot is shown at each level of occlusion defined in IRUO. These occlusion levels are exactly the same as those defined in the OVIS dataset.

There exists an official, author-defined train-test split for OVIS; however, labels for the test partition are unavailable due to the ongoing OVIS competition (\cite{qi_occluded_2022}). Therefore, in this work we partitioned the official OVIS training set into training and test sets for IRUO. We then remove images in the \textit{boat} and \textit{vehicle} classes, as there are very few examples of each at occlusion level 0, leaving 23 remaining classes of objects. For each of these 23 classes of objects, we split the set of \textit{videos} (not images) such that approximately 70 percent of instances of the class appear in the training data and 30 percent appear in the test data. We ensure that, per-class, no instances of objects in the training set appear in the test set, to prevent positively-biased estimates of model accuracy.

\begin{figure}[ht]
\begin{center}
\includegraphics[width=0.92\linewidth]{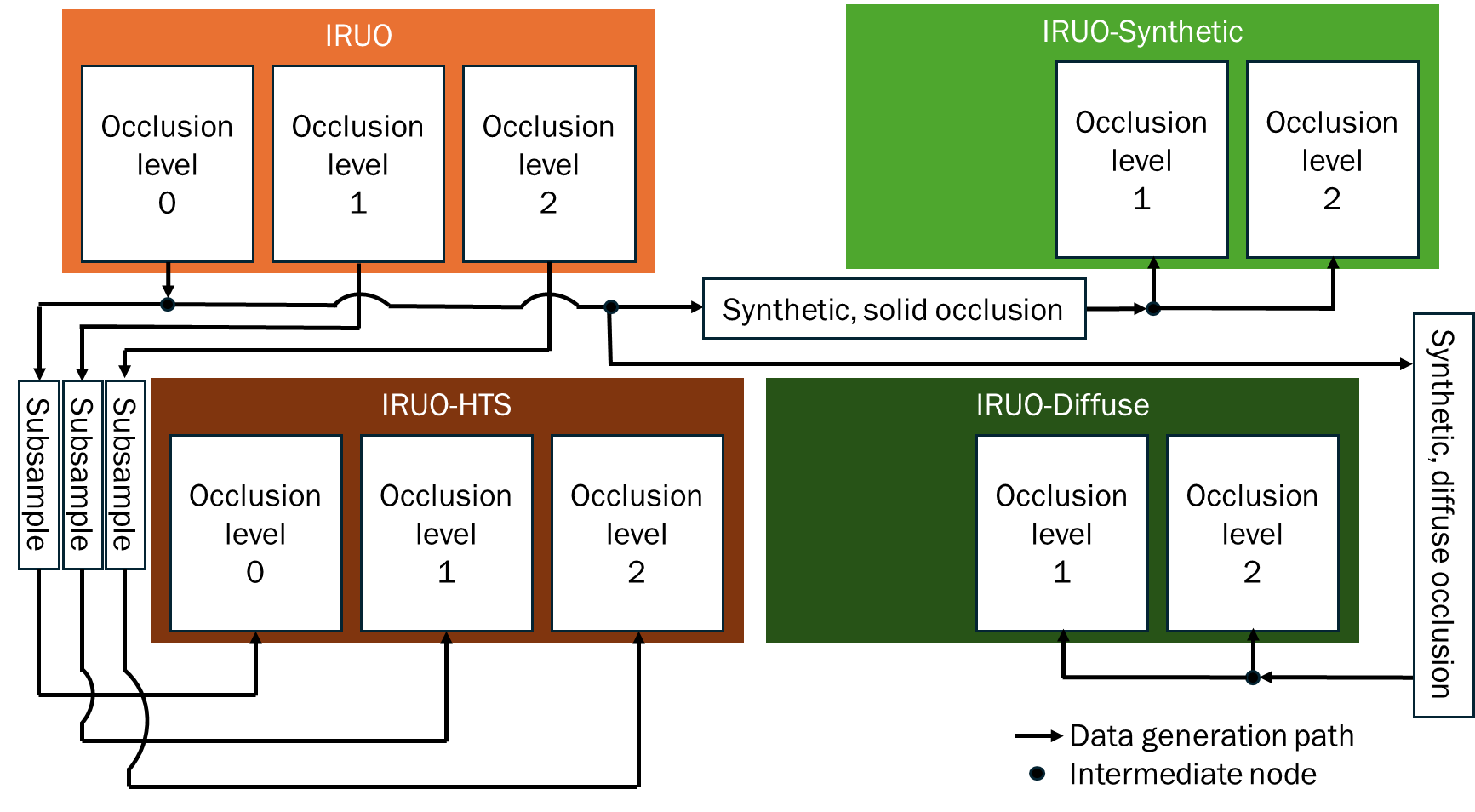}
\caption{Diagram showing the data generation process of all dataset partitions included in IRUO. A description of each partition is found in Table \ref{tab:dataset_details}}
\label{fig:dataset_generations}
\end{center}
\end{figure}

From the base IRUO dataset, we derive three additional datasets: IRUO-HTS, IRUO-Synthetic, and IRUO-Diffuse. IRUO-HTS refers to the subset of IRUO utilized to estimate human recognition accuracy under occlusion in \ref{sec:human_study}. We use IRUO-Synthetic to examine whether synthetic occlusions are a good proxy for real-world occlusions, in Sec. \ref{sec:synth}. We utilize IRUO-Diffuse to examine the relative accuracy of human and DNN-based models to diffuse occluders, a particular type of occlusion, in Sec. \ref{sec:diffuse}. We provide further details about the derivation of each of these datasets in the sections associated with each one; however, a summary is provided in Fig. \ref{fig:dataset_generations}. We provide a quantitative summary of all four IRUO partitions in Table \ref{tab:dataset_details}. 

\begin{table*}[htb]
    \centering
    \begin{tabular}{p{0.2\linewidth}|p{0.1\linewidth}p{0.1\linewidth}p{0.5\linewidth}}
    Name & Training images & Test images & Description \\
         \hline
    IRUO-Base & 61.4k & 26.4k & Original partition of IRUO, created using object instances from OVIS (\cite{qi_occluded_2022}). \\
    IRUO-HTS & 0 & 552 & Subsampled \textit{human testing subset} version of IRUO, with blur and small-target filters, 8 images per class per occlusion level. \\
    IRUO-Synthetic & 64.0k & 24.5k & Dataset partition generated from unoccluded images in IRUO. Contains synthetic occluders such as noise, textures, solid boxes, and superimposed objects. \\
    IRUO-Diffuse & 114.9k & 68.5k & Dataset partition of IRUO generated from unoccluded images in IRUO. Contains synthetic occluders such as regular lines, grids, diagonal grids, and small squares at 1, 2, 4, 8, and 16px size. \\
    \end{tabular}
    \caption{Details of IRUO dataset partitions used in this study}
    \label{tab:dataset_details}
\end{table*}

\section{Human Study}\label{sec:human_design}

In this section we describe the process for estimating human-level recognition accuracy under varying levels of occlusion. To do this, we recruited \textbf{20} human subjects to classify a curated subset of images from the testing set of our IRUO dataset, termed the Human Testing Subset (IRUO-HTS). Below we describe how we designed and utilized IRUO-HTS, as well as the testing protocol used to collect the subjects' predictions on IRUO-HTS. 

\subsection{The Human Testing Subset}
\label{sec:human_filtering}
To mitigate the risks of subject fatigue, each human subject was only asked to classify 200 test images which was chosen to require about 1 hour to complete using our testing protocol\footnote{The human subjects research described in this publication is pursuant to Duke University Campus IRB protocol \#2023-0336.}. These 200 images were sampled from a curated subset of 552 images from the full IRUO testing dataset, which we term the "Human Testing Subset" (IRUO-HTS). To ensure that IRUO-HTS was representative of the IRUO test dataset, we randomly sampled 8 images for each of the 23 target object classes, and did so for each of 3 occlusion levels, resulting in a total of 552 images. We sub-sample images from IRUO-HTS (e.g., instead of the full IRUO dataset) to ensure that multiple subjects classify each image, on average. This has the drawback of limiting the total number of unique images in the test set, while allowing us to identify and remove human subjects that consistently performed worse than others (i.e., outliers). Because each test image varies in its difficulty, some subjects may perform worse simply because their subset of testing images is more difficult. By enforcing some overlap in test images, we can compare the accuracy of each subject to several other subjects over the same images. Using our sampling procedure on IRUO-HTS, and our total of 20 human subjects, each image was classified by about 7 subjects on average. 

\begin{figure}[htb]
 \begin{center}
\begin{tabular}{cc}
\includegraphics[width=0.45\linewidth]{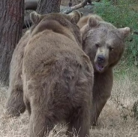}&
\includegraphics[width=0.45\linewidth]{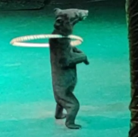}\\
(a)&(b)\\
\includegraphics[width=0.45\linewidth]{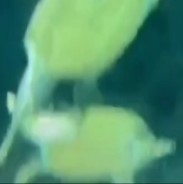}&
\includegraphics[width=0.45\linewidth]{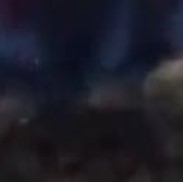}\\
(c)&(d)
\\
\end{tabular}
 \end{center}
\caption{Top row, (a), (b): sample images kept by blur filter algorithm; bottom row, (c), (d): sample images rejected by blur filter algorithm. The blur filter parameters used are a minimum Laplacian variance of 20 and minimum image size of 10,000 pixels}
\label{fig:blur_samples}
\end{figure}

We choose to include 8 samples per class per occlusion level to balance variation between humans and variation between images, i.e., too many images may result in few human observers seeing each image, and too few images may risk not accurately reflecting human accuracy. We additionally isolate the effect of occlusion from other factors, including small imagery and blur, by applying two additional selection criteria for images in this subset. We approximate blur using the variance of the Laplacian operator, as in \cite{bansal_blur_2016}, and we approximate image size by using the number of cropped pixels on image targets. We choose values for minimum Laplacian variance and image size that give us approximately 90\% accuracy on top models at occlusion level 0; in our experiments, these values are 20 and 10,000, respectively. Images filtered using this blur algorithm are shown in Fig. \ref{fig:blur_samples}.

\subsection{Testing Protocol}
\label{sec:human_labeling}

\begin{figure}[htb]
 \begin{center}
\begin{tabular}{c}
\includegraphics[width=0.9\linewidth]{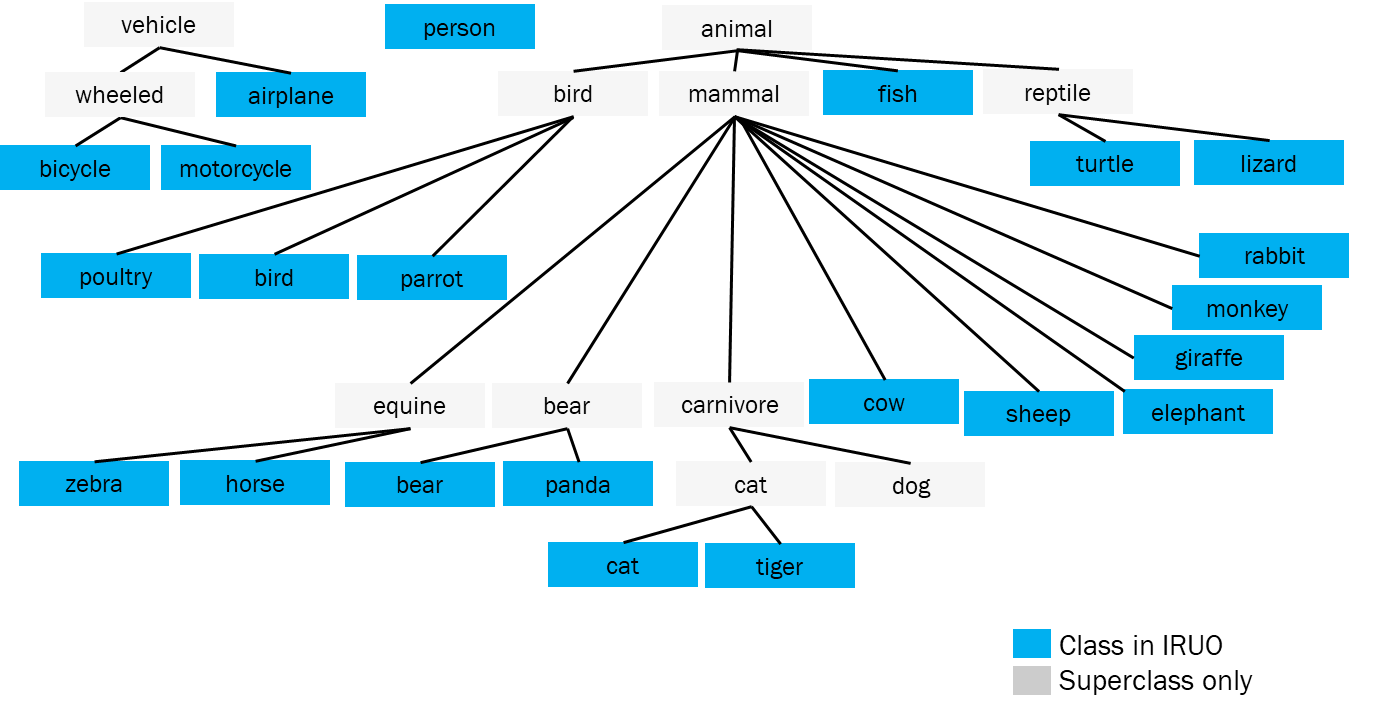}
\end{tabular}
 \end{center}
\caption{Hierarchy of classes in IRUO. In Sec. \ref{sec:human_study} and Fig. \ref{fig:top_humans}, results are reported according to \textit{classification level}, or the tree depth reported for each class on this tree; e.g., the class "zebra" is located at level 4, and the class "fish" is located at level 2. Note that classes that appear to be subclasses of themselves refer to classes in which there is an "other" category, e.g., the subset "cat" refers to superset "cat" objects that are not "tiger" objects}
\label{fig:hierarchy}
\end{figure}

We employ a method of human labeling known as \textit{Scalable Multi-Label Annotation}, first used in \cite{deng_scalable_2014}. This method of data collection attempts to minimize human error due to factors other than strict identification of classes. As described in detail in Sec. \ref{sec:human_labeling_procedure}, this collection procedure is designed to address the difference between model memorization of classes and humans' comparatively limited memory by relying on hierarchical prompts. We use the hierarchy in Fig. \ref{fig:hierarchy}, derived from WordNet, a commonly-used lexical database of the English language (\cite{miller_wordnet_1995}). This hierarchy gives all classes in IRUO a hierarchy "level," i.e., the number of clicks required to obtain this class in the human study program. For example, the class "fish" does not appear at level 1 but appears at levels 2-5, and the class "dog" does not appear at levels 1-3 but appears at levels 4-5. 

To ensure that human observers select the correct object within an image (e.g., in cases where there are multiple classifiable objects in frame), we additionally ask humans to label which object they are selecting by clicking a point on the image. The correct object is usually centered on the image, so we place a cross directly at the center of each image for human observers. This guides human observers toward the correct object, but it does not give away information about which object to classify that is not available to models; this cross does not move throughout data collection.

\section{Experimental Results}\label{sec:benchmark}

Each subsection below aims at addressing a specific scientific question, given by its title. In each case we report experimental results with our IRUO dataset that address the given question. All models in our experiments utilize encoders that have already been pre-trained on the ImageNet-1k dataset (\cite{deng_imagenet_2009}). We consider a large number of models that combine different architectures (e.g., convolutional, transformer, etc) and augmentation strategies); we fine-tune each model on the IRUO dataset. Following previous work suggesting that training on occluded imagery is not beneficial for models (\cite{kortylewski_compositional_2020}, \cite{zhu_robustness_2019}) we fine-tune the models only on unoccluded imagery. We independently optimize several key hyperparameters (e.g., learning rates, momentum) for each model using a grid search to find highest overall accuracy on IRUO. Full details of the hyperparameter grid search, optimized hyperparameter choices (including hyperparameters unique to models designed for occlusion), stopping criteria, and other training details can be found in the Appendix.

\subsection{Which models are most accurate under occlusion?}

\label{sec:results}

\begin{figure}[t]
 \begin{center}
\begin{tabular}{c}
\includegraphics[width=0.99\linewidth]{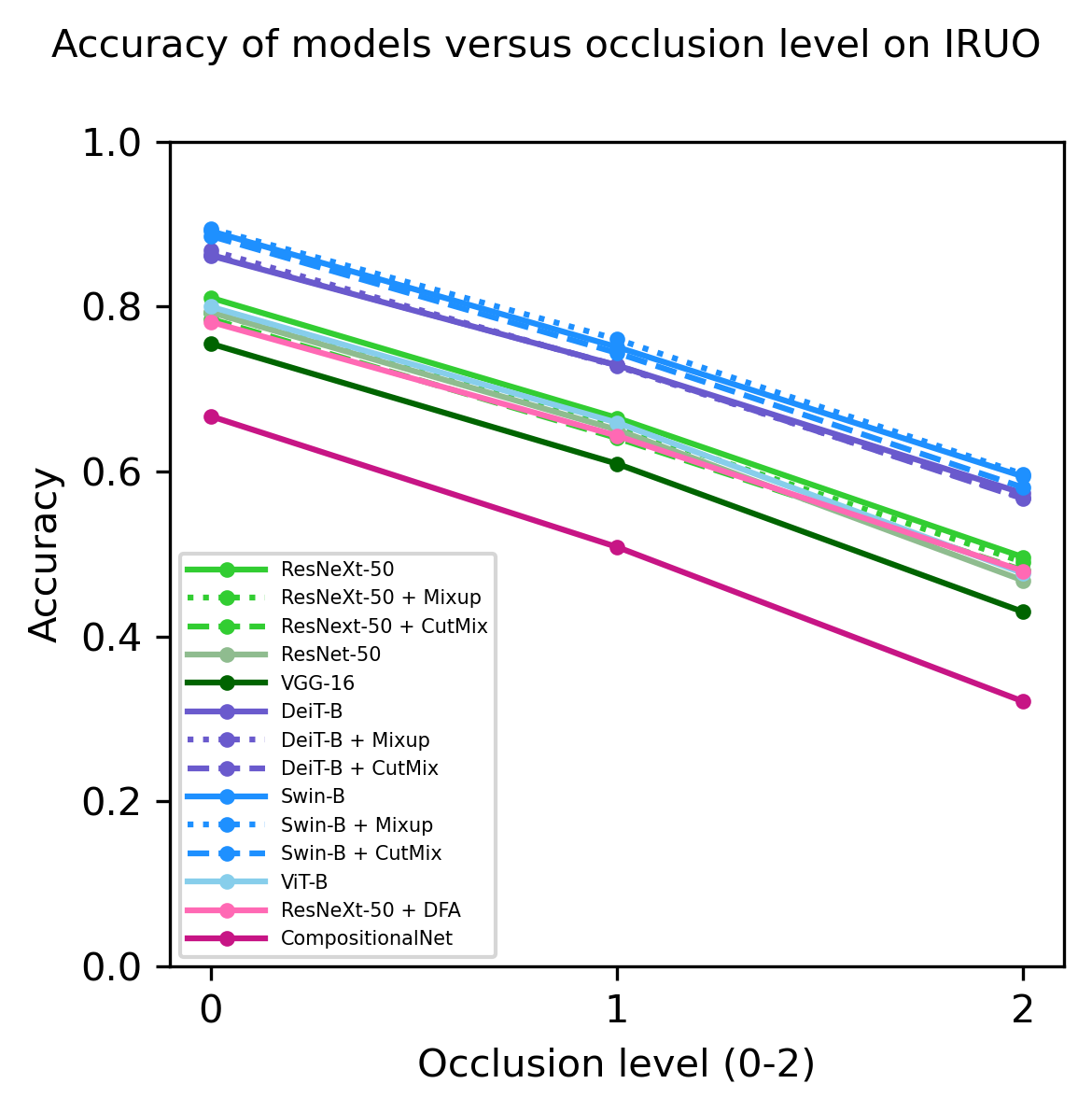}
\end{tabular}
 \end{center}
\caption{Accuracy per level of occlusion on the partition of the IRUO-Base dataset containing real occlusion, measured as proportion of correct top-1 classifications for each model at each occlusion level. The x-labels 0, 1, and 2 correspond to the occlusion levels in Sec. \ref{sec:dataset}. Convolutional models are highlighted in green, Vision Transformer-based models are highlighted in blue, and models explicitly designed for occlusion robustness are highlighted in pink. The line style (solid, dot, dash) indicate augmentations used for selected models. Deep Feature Augmentation (\cite{cen_deep_2021}) is abbreviated as DFA for clarity}
\label{fig:our_accuracy}
\end{figure}

To address this question, we report in Fig. \ref{fig:our_accuracy}  the accuracy of the models and augmentations described in Sec. \ref{sec:models} and Sec. \ref{subsec:augmentations}, respectively, under varying levels of occlusion. The results indicate that Transformer-based models (blue and purple lines) perform best, by a large margin, across all levels of occlusion, including Level 0 (no occlusion). The Swin model (blue lines) achieves the best overall performance among Transformers, making it the best-performing model in our benchmark. The convolutional models (green lines in all but one case) achieve the next best accuracy. Among the convolutional models we see that ResNeXt performs best, followed by ResNet, and then VGG. Similar to transformer-based models, the relative rank of the convolutional models (compared to others in the benchmark) is similar across all levels of occlusion. Surprisingly, the worst-performing model is the CompositionalNet. We observe mixed and minimal impact of the two augmentation schemes that we considered: Mixup and CutMix. The best-performing model is the Swin model with  Mixup; however, the performance gain over Swin alone is small, and the impact of Mixup on other models is more modest, or even sometimes negative. 

\begin{figure}[ht]
 \begin{center}
\begin{tabular}{cc}
\includegraphics[width=0.47\linewidth]{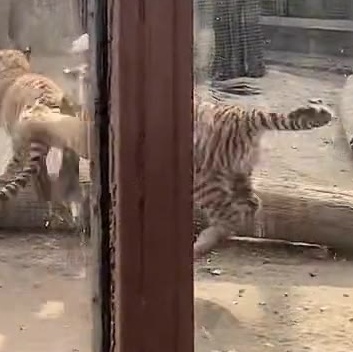}&
\includegraphics[width=0.47\linewidth]{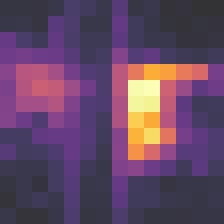}\\
(a)&(b)\\
\includegraphics[width=0.47\linewidth]{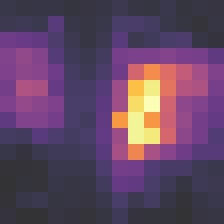}&
\includegraphics[width=0.47\linewidth]{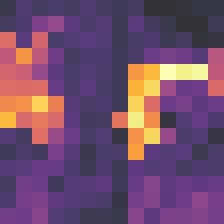}\\
(c)&(d)
\end{tabular}
 \end{center}
\caption{Sample image of a tiger (a) at occlusion level 2 with corresponding self-attention maps returned by the ViT-B model trained on IRUO at the (b) third, (c) seventh, and (d) eleventh transformer stages. Self-attention maps are summed over all points on target and all heads for each location in each layer displayed. ViT and other transformer models demonstrate the ability to attend to relevant objects and ignore occluders in certain cases. Lighter colors (e.g., yellow) denote higher values for attention}
\label{fig:vit_attention_maps}
\end{figure}

Our findings expand upon and in some cases modify findings of prior work. Recent work (\cite{naseer_intriguing_2021}) found that transformers outperform convolutional models in the presence of occlusion; our work corroborates this finding on a more comprehensive and controlled experimental setting. The authors in \cite{naseer_intriguing_2021} hypothesized that this increase in accuracy under occlusion is due to the ability of receptive fields that result from self-attention layers to adapt to occlusion and attend to unoccluded data while down-weighting the relative importance of occluded data. We corroborate these results qualitatively in Fig. \ref{fig:vit_attention_maps}; an image of a highly-occluded tiger is shown in Fig. \ref{fig:vit_attention_maps}(a) alongside corresponding attention maps generated by ViT. To calculate the attention masks, we select all patches with object pixels, and we sum the values of attention at all other locations for all attention heads. As shown in Fig. \ref{fig:vit_attention_maps}, objects are clearly separated from background and occluders, and parts of the tiger in frame attend to each other. This suggests that Vision Transformer-based models can localize relevant objects and ignore occluders. Recent work reported that the CompositionalNet model outperforms convolutional models (\cite{kortylewski_compositional_2020}); however, we find here that all DNN-based models outperform it. We hypothesize that this discrepancy is due to the significantly greater complexity of the IRUO dataset here (e.g., in terms of scene and target class diversity) compared to the \textit{VehicleOcclusion} dataset utilized in \cite{kortylewski_compositional_2020}. Lastly, Deep Feature Augmentation (\cite{cen_deep_2021}) demonstrated higher accuracy under occlusion however we find more mixed results. This possibly reflects a difference in our implementation compared to the original; we do not use the same synthetic occlusions in training and testing, as is done in \cite{cen_deep_2021}.

As noted above, our results indicate that the accuracy of every model degrades at approximately the same rate as the level of occlusion increases. Consequently, the relative accuracy of a given model in the presence of occlusion (i.e., compared to other models) is well-predicted by its relative accuracy on unoccluded imagery. For example, Transformer-based models (blue lines) perform best on unoccluded imagery, and we observe that they likewise also perform best on all other levels of occlusion. Interestingly, these results imply that none of the models considered in our benchmark are advantageous \textit{specifically for occlusion}: i.e., any advantages/disadvantages of each model in the presence of occlusion are also similarly advantageous for unoccluded imagery, and vice versa. A corollary of this finding is that one can improve accuracy under occlusion by improving accuracy on unoccluded imagery. Although this pattern holds for a large number of models, examined on a diverse dataset, this observation is empirical and therefore its generality may ultimately be limited. 

\subsection{Is synthetic occlusion a good proxy for real occlusion?}
\label{sec:synth}

\begin{figure}[htb]
\begin{center}
\begin{tabular}{cccc}
\includegraphics[width=0.4\linewidth]{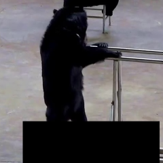}&
\includegraphics[width=0.4\linewidth]{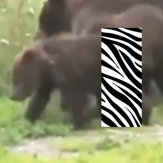}\\
(a)&(b)\\
\includegraphics[width=0.4\linewidth]{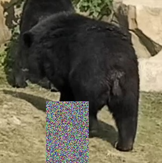}&
\includegraphics[width=0.4\linewidth]{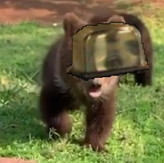}\\
(c)&(d)
\end{tabular}
 \end{center}
\caption{Images showing various types of artificial occlusion used in our benchmark. In (a) black mask occlusion, (b) texture occlusion, (c) noise occlusion, and (d) superimposed object occlusion using objects from COCO (\cite{lin_microsoft_2015}), all based on artificial occlusions used in \cite{tang_recurrent_2018,kortylewski_compositional_2020,naseer_intriguing_2021,zhu_robustness_2019}}
\label{fig:our_artificial_occlusions}
\end{figure}

In this section we investigate whether synthetic occlusions are an accurate proxy for real-world occlusions, when estimating the \textit{relative} accuracy of competing models. In other words, we ask whether the evaluation of model accuracy using synthetic occlusions would lead to the same rank-ordering of models as real-world occlusion. To address this question, we evaluate model accuracy using two datasets: one with real-world occlusions, and one with synthetic occlusions. If the rank-orderings produced by each dataset are highly dissimilar, then we conclude that synthetic occlusions are not a reasonable proxy for real-world occlusions. We use the IRUO-Base and IRUO-Synthetic partitions of the IRUO dataset to rank-order models on real-world and synthetic occlusions, respectively. The construction of the IRUO-Base dataset is detailed in Sec. \ref{sec:dataset}. The IRUO-Synthetic dataset was created by adding artificial occlusion to all of the images in IRUO-Base that do not contain occlusion. To create synthetic occlusions, we replicate several strategies utilized in recent prior work (e.g., \cite{kortylewski_compositional_2020, zhu_robustness_2019, naseer_intriguing_2021, chen_transmix_2022}), which share the same basic strategy of placing (i.e., inpainting) occluding objects at pseudo-random locations over the target object in the imagery. The primary difference among prior approaches is the design of the occluding objects. We consider several previously explored occluder designs: white boxes, black boxes, zebra-print texture boxes, uniformly-distributed noise boxes, and objects from the COCO dataset (\cite{lin_microsoft_2015}). Examples of the synthetic occlusions are presented in Fig. \ref{fig:our_artificial_occlusions}. Full details of our occluder designs and the occluder placement strategy can be found in the Appendix. 

\begin{figure*}[ht]
\begin{center}
\begin{tabular}{cc}
\includegraphics[width=0.48\linewidth]{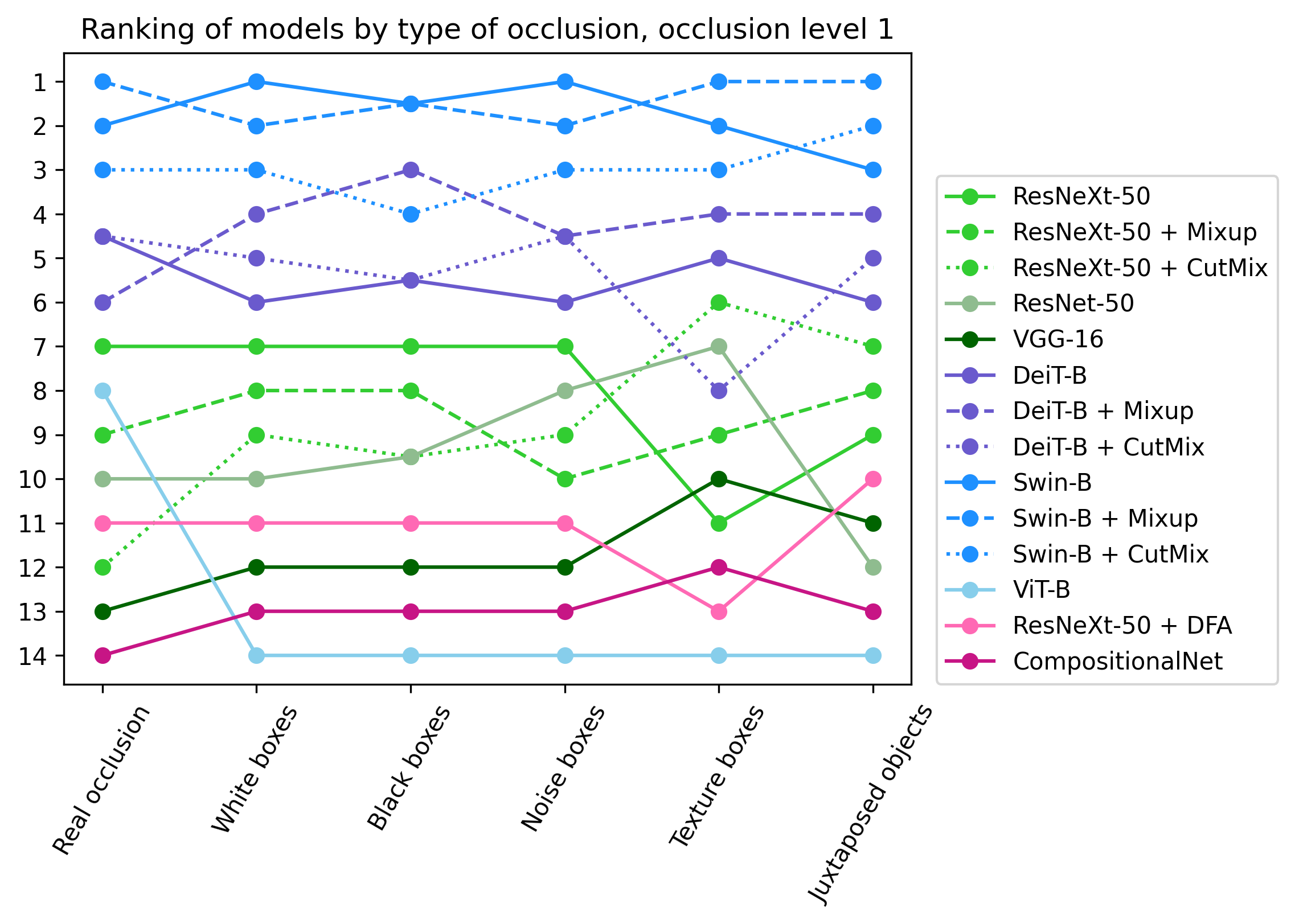}&
\includegraphics[width=0.48\linewidth]{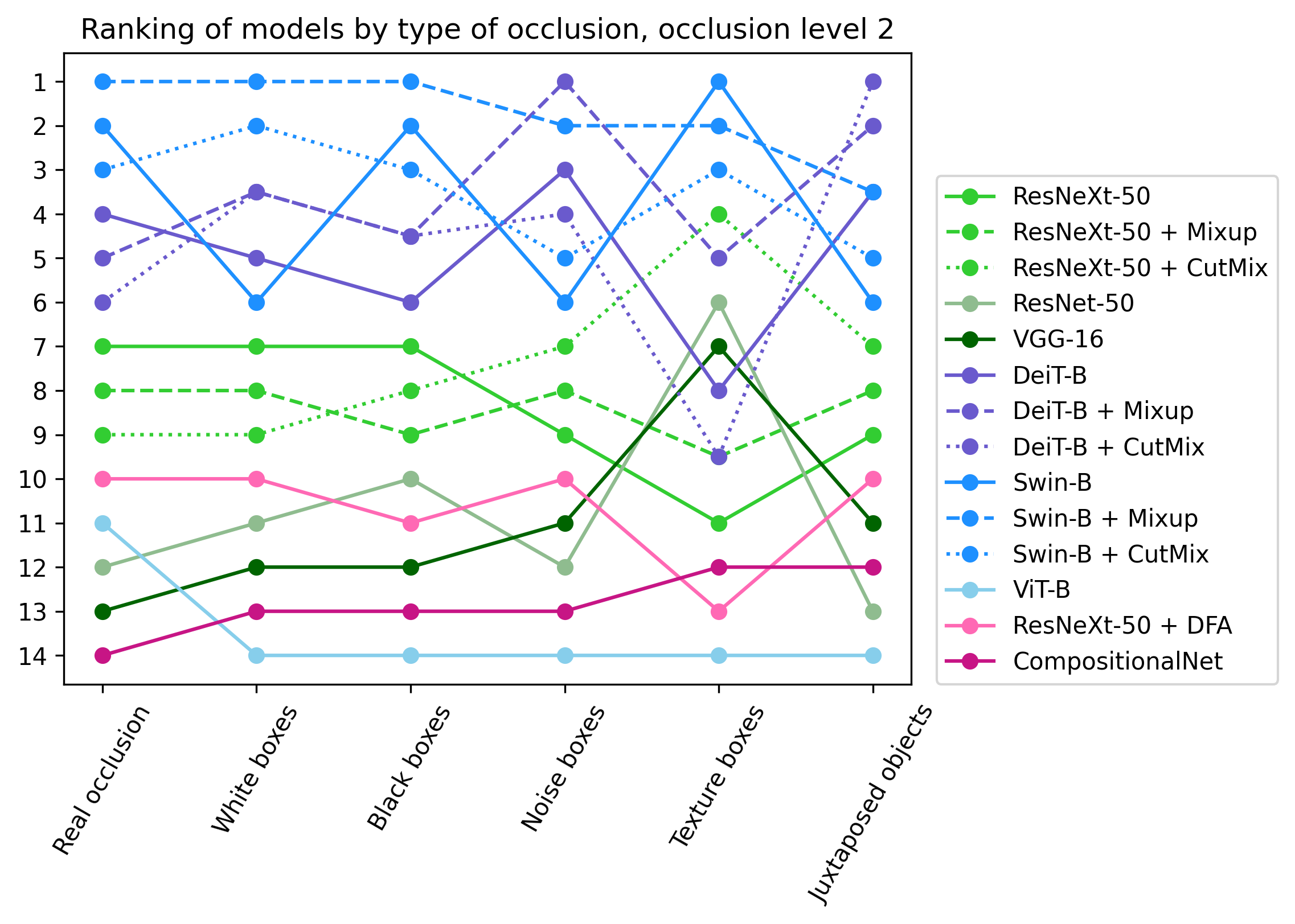}\\
(a)&(b)
\end{tabular}
\end{center}
\caption{Rankings of models across occlusion types at each of (a) occlusion level 1 and (b) occlusion level 2}
\label{fig:bump_chart}
\end{figure*}

\textbf{Qualitative Assessment.} First we qualitatively evaluate the agreement of the rankings provided by real and synthetic occlusion. In Fig. \ref{fig:bump_chart}(a), we present a line plot showing the ranking of each model as we change the occlusion type for occlusion level 1. The results indicate that no pair of occlusion types produces the same ranking of models; however, the the rankings are always very similar. For example, models often don't change ranking, and when they do, it is only by one or two positions. The variability of ranking is even more limited if we only consider changes in architecture (i.e., crossings between solid lines). In terms of similarity to real occlusion, it appears that solid boxes (white or black boxes), or juxtaposed objects, provide the most similar rankings. In Fig. \ref{fig:bump_chart}(b), we present the rankings of models for occlusion level 2, where we observe somewhat lower consistency in the rankings; however, the conclusions are otherwise the same. 

Our qualitative results therefore suggest that synthetic occlusions often provide good approximate \textit{rankings} of model accuracy under occlusion. The rankings provided by synthetic occlusion are not identical to real-world occlusions, however; therefore, synthetic occlusions will be a better proxy for large differences in model accuracy. It is also noteworthy that the variations in rankings by each occlusion type may not be due to systematic differences among occlusion types, and instead may be due to random variations in the data; i.e., testing with a bootstrap sample of the test imagery of one occlusion type may produce different rankings, as well. 

\textbf{Statistical Assessment.} To provide a more precise notion of similarity among the rankings, we utilize the Friedman $Q$-value test  (\cite{friedman_use_1937}), summarized in Sec. \ref{sec:stats}. We treat the benchmark models as ranked "objects" in the Friedman test and each of occlusion type as "judges" that rank the objects. Using the rank-ordering of models provided by each occlusion type, we can then calculate the Friendman $Q$ statistic to test the null hypothesis, which supposes that the rankings provided by the occlusion types are random. Therefore, if we reject the null hypothesis, particularly with very large values of $Q$, this suggests there is high similarity in the rankings provided by each occlusion. We perform the Friedman test separately for each of two occlusion levels: level 1 (0-50\%) and level 2 ($>$50\%). For each occlusion level, we treat each of the six types of occlusion (real occluders, white boxes, black boxes, noise boxes, texture boxes, and synthetically added objects) as a set of six judges. The calculated Friedman $Q$-value follows a chi-squared null distribution for either a large number of models ($k>4$) or types of occlusion ($n>15$) (\cite{friedman_use_1937}); as our number of models, $k$, is 14, the former condition is met in our study, and therefore our calculated $Q$ value is approximately $\chi^2_{13}$-distributed. Full calculations for this statistic can be found in the Appendix. 

In Table \ref{tab:rank_summary}, we present the calculated values of $Q$ for occlusion levels 1 and 2. Based on the very large values of $Q$ for both levels of occlusion and the resulting small p-values (far less than 0.05), we reject our null hypothesis, determining that there is a consistent rank order of models at each occlusion level, regardless of the specific type of occlusion. This implies that, for occlusion levels 1 (0-50\%) and 2 (\textgreater50\%), any type of occlusion tested is appropriate for determining the effect of occluded objects on classification accuracy. A summary of these values may be found in Table \ref{tab:rank_summary}, and full rankings and calculations of these values may be found in Table \ref{tab:full_rank}. 

\begin{table}[h!]
    \centering
    \begin{tabular}{c|cc}
         \hline
         Occlusion level & $Q$ & p-value\\
         \hline
         1 (0-50\%) & 71.0 & $5.0\times10^{-10}$\\
         2 (\textgreater50\%)& 65.5 & $5.3\times10^{-9}$\\
         \hline
    \end{tabular}
    \caption{Calculated values of $Q$ and corresponding p-values assuming a null distribution with $k$ degrees of freedom, following the Friedman $Q$ procedure in Sec. \ref{sec:stats}, based on comparisons between the IRUO and IRUO-Synthetic datasets}
    \label{tab:rank_summary}
\end{table}

\subsection{How occlusion-robust are models compared to humans?}
\label{sec:human_study}

In this section we investigate how occlusion-robust DNNs are to occlusion. We determine robustness by comparing the accuracy of DNNs to the accuracy of a hypothetical optimal model, which we approximate using human recognition accuracy under occlusion (see Sec. \ref{sec:human_design}). 

\begin{figure}[htb]
 \begin{center}
\begin{tabular}{c}
\includegraphics[width=0.9\linewidth]{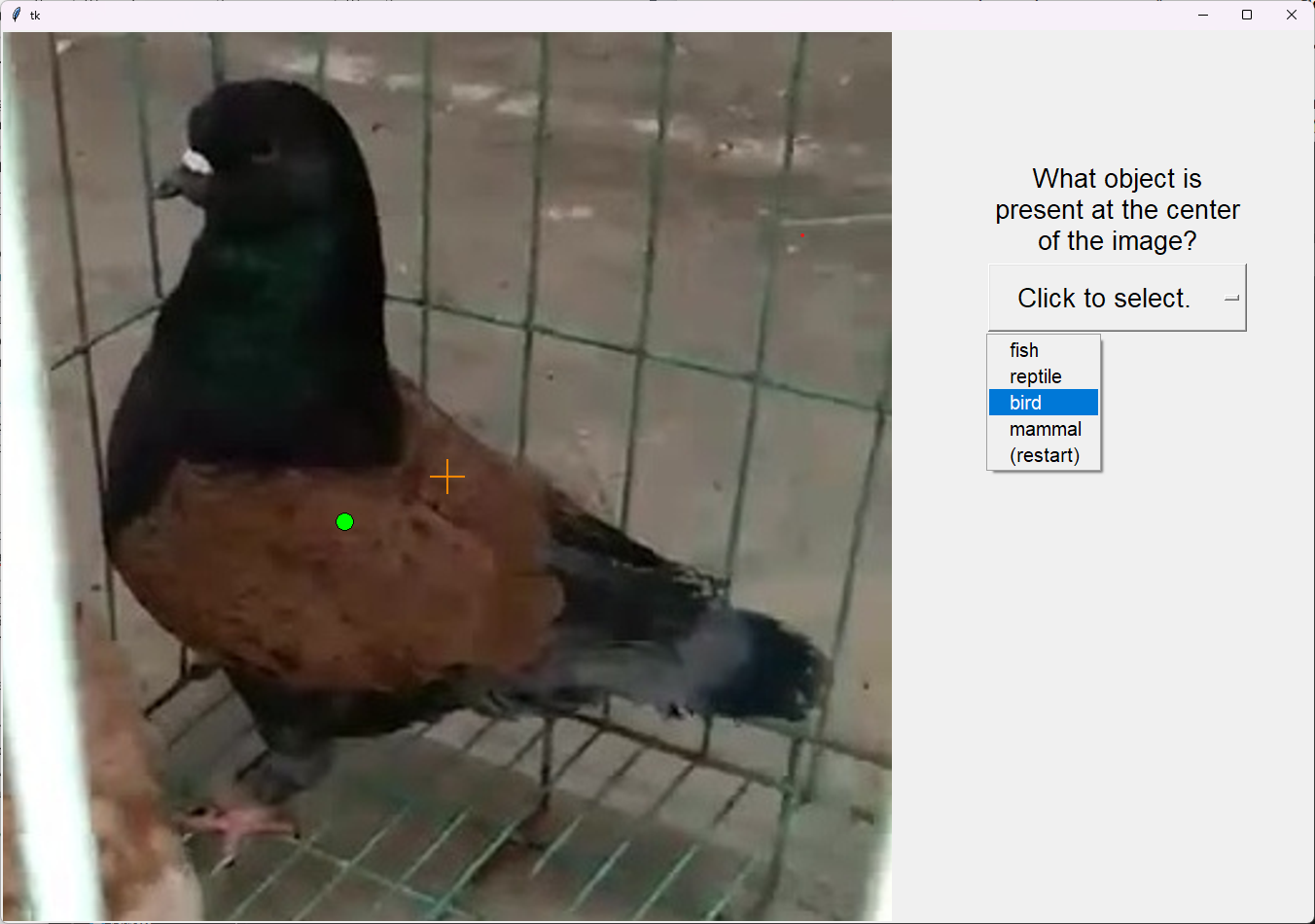}
\end{tabular}
 \end{center}
\caption{Sample interface from our human study collection program. Human subjects are asked to label images as one of 23 classes listed in the IRUO dataset using an interface following guidelines described in \cite{deng_scalable_2014}}
\label{fig:sample_program}
\end{figure}

\textbf{Human Study Design.}  To assess human accuracy, we use the \textit{IRUO-HTS} partition of our IRUO dataset, the creation of which is described in Sec. \ref{sec:human_filtering}. We ask each of 20 people, or \textit{observers}, to label 200 randomly-selected images from this partition (of size 552 images). We follow the method in Sec. \ref{sec:human_labeling}, asking respondents for image labels at each level of the hierarchy in Fig. \ref{fig:hierarchy} until a specific class label belonging to IRUO is selected. A sample step of the program interface used to collect human responses is shown in Fig. \ref{fig:sample_program}. We then estimate the human accuracy for the $j^{th}$ image by 
\begin{equation}
q_{j} = \frac{1}{N_{j}} \sum_{i \in S_{j}} \mathbbm{1} (\hat{t}_{i,j} = t_{j})
\label{eq:acc_per_image}
\end{equation}
where $S_{j}$ is the set of observers that labeled the $j^{th}$ image, $\hat{t}_{i,j}$ is the label assigned to the $j^{th}$ image by the $i^{th}$ observer, and $t_{j}$ is the ground truth label for the $j^{th}$ image. Then, to calculate average human accuracy, we take the average of $q_{j}$ values across all images in the IRUO-HTS partition. 

\textbf{Identification of Outlier Observers.}  Some human observers may consistently obtain lower recognition accuracy than others (e.g., due to fatigue, or misunderstanding directions), implying that they may negatively bias our estimate of optimal recognition. To mitigate this potential problem, we remove human observers from our estimate that consistently perform worse than other observers \textit{when examined on the same imagery}. It is crucial to compare humans (or any recognition model) on a shared set of imagery because classification difficulty varies greatly between images. Therefore we compute a "normalized accuracy" for each observer, which is their accuracy relative to other observers that rated the same imagery. Once we obtain the normalized accuracy for each observer, we use Tukey's Rule (see Sec. \ref{sec:stats}) to remove outliers: i.e., observers that consistently perform worse than others over the same imagery.

We define the normalized accuracy of the $i^{th}$ observer, as $a^{A}_{i} = a_{i} - \bar{a}_{i}$, where $a_{i}$ is the proportion of top-1 accuracy of the $i^{th}$ observer, and $\bar{a}_{i}$ is an estimate of the average top-1 accuracy of all observers on the subset of 200 images shown to the $i^{th}$ observer. We compute $\bar{a}_{i}$ as 
\begin{equation}
\bar{a}_{i} = \frac{1}{200} \sum_{j\in S_i} {q_{j}}
\label{eq:human_average}
\end{equation}
where $S_i$ is the subset of 200 images shown to the $i^{th}$ observer, and $q_{j}$ is the human accuracy for image $j$, as defined in Eq. \ref{eq:acc_per_image}. We then apply Tukey's rule to identify and remove outliers from the set of normalized accuracy values for our 20 observers, denoted $S^A = {\{s^A_{i}\}}_{i=1}^{20}$. This resulted in the detection and removal of the two lowest-performing participants from the study.

\begin{table*}[ht]
    \centering
    \resizebox{\linewidth}{!}{%
    \begin{tabular}{ccccc}
         \hline
         \multirow{2}{*}{Model} & \multirow{2}{*}{Augmentation} & \multicolumn{3}{c}{Accuracy: occlusion level} \\
         & & 0 & 1 & 2 \\
        \hline
         VGG-16 (\cite{simonyan_very_2015}) & -  & 0.755 & 0.674 & 0.554 \\
         CompositionalNet (\cite{kortylewski_compositional_2020})  & - & 0.679 & 0.543 & 0.440\\
         ViT-B (\cite{dosovitskiy_image_2021}) & - & 0.674 & 0.663 & 0.614 \\
         ResNet-50 (\cite{he_deep_2016}) & - & 0.761 & 0.696 & 0.587 \\
         \hline
         \multirow{4}{*}{ResNeXt-50 (\cite{xie_aggregated_2017})} & - & 0.799 & 0.690 & 0.620 \\
         & Mixup (\cite{zhang_mixup_2018}) & 0.772 & 0.658 & 0.576 \\
         & CutMix (\cite{yun_cutmix_2019}) & 0.772 & 0.636 & 0.582 \\
         & Deep Feature (\cite{cen_deep_2021})& 0.734 & 0.668 & 0.576 \\
         \hline
         \multirow{3}{*}{DeiT-B (\cite{touvron_training_2021})} & - & 0.842 & 0.783 & 0.723 \\
         & Mixup (\cite{zhang_mixup_2018}) & 0.880 & 0.821 & 0.723 \\
         & CutMix (\cite{yun_cutmix_2019})) & 0.886 & 0.832 & 0.734 \\
         \hline
         \multirow{3}{*}{Swin-B (\cite{liu_swin_2021})} & - & 0.918 & 0.826 & 0.744 \\
         & Mixup (\cite{zhang_mixup_2018}) & \textbf{0.940} & \textbf{0.837} & \textbf{0.772} \\
         & CutMix (\cite{yun_cutmix_2019}) & 0.929 & 0.815 & 0.755 \\
         \hline
         Human (average, outliers removed) & - & 0.947 & 0.862 & 0.800  \\
         \hline
    \end{tabular}}
    \caption{Results of models and humans on IRUO-HTS. Outliers are removed according to the procedure in Sec. \ref{sec:stats}. The entry labeled "Human (top responses)" indicates the average score of the top nine respondents}
    \label{tab:human_model_accuracy}
    
\end{table*}

\begin{figure*}[ht]
 \begin{center}
\begin{tabular}{c}
\includegraphics[width=0.97\linewidth]{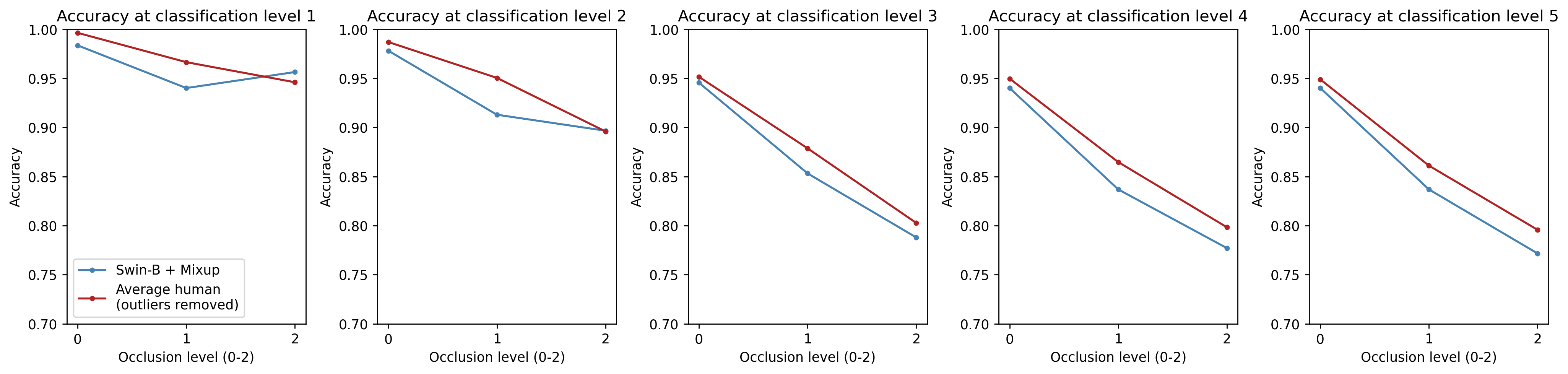}
\end{tabular}
 \end{center}
\caption{Accuracy per level of occlusion on the partition of IRUO containing real occlusion, measured as proportion of correct top-1 classifications for each model at each occlusion level, considering \textit{only} images seen by the highest-scoring human respondents. Occlusion level 0 corresponds to no occlusion, 1 corresponds to 0-50\% occlusion, and 2 corresponds to \textgreater50\% occlusion. Classification level is defined as the tree depth of each IRUO class listed in Fig. \ref{fig:hierarchy}. For this set of humans, outliers are removed according to the procedure in Sec. \ref{sec:stats}. Full results, including all models and augmentations, evaluated on the IRUO subset (at classification level 5) shown to humans, are shown in Table \ref{tab:human_model_accuracy}}
\label{fig:top_humans}
\end{figure*}

\textbf{Results.} The results are presented in Table \ref{tab:human_model_accuracy}, where we report the accuracy of all benchmark models, as well as an average of the human observers (with outliers removed), on the IRUO-HTS dataset. Among the benchmark models, the results are similar to those of Sec. \ref{sec:results} with the larger IRUO dataset, and the overall conclusions are the same. In particular, the Swin model with Mixup (denoted Swin-Mixup here) achieves the best accuracy among DNN-based models across all levels of occlusion, including none. Therefore we treat Swin-Mixup as a representative of state-of-the-art DNN-based models for both occluded and non-occluded tasks, and focus our discussion on comparing Swin-Mixup against human-level accuracy. The results indicate that the Swin-Mixup model obtains nearly the same accuracy as humans on unoccluded imagery, which is consistent with recent comparisons human and DNN image recognition accuracy (e.g., in \cite{liu_swin_2021}, \cite{touvron_training_2021}). 

However, when occlusions are present in the imagery, we find that humans outperform Swin-B with Mixup (and therefore all of our DNN-based models), indicating that DNN-based models are not yet fully occlusion-robust. However, surprisingly, we find that the performance gap between Swin-Mixup and humans is relatively modest: approximately $3\%$ for Level 1, and $3.5\%$ for Level 2. The performance gaps are somewhat larger for other variants of Swin (e.g., without augmentation, or with CutMix), however, the gaps are still relatively small. The results therefore suggest that modeling improvements are possible; however, the potential benefits are relatively modest when considering state-of-the-art models, such as vision transformers. Crucially, these results provide some insight into precisely what level of improvements can be expected when investing in further methodological improvements, and our IRUO dataset provides a benchmark to assess the effectiveness of new models.

One potential explanation for the relatively small human-to-DNN performance gap may be that the number of classes in IRUO is large, and fairly nuanced: e.g., with classes as specific as \textit{bird}, \textit{parrot}, and \textit{poultry}. Due to their limited knowledge and/or memory capacity, humans may be unable to remember the precise visual differences between some target classes, leading to underestimation of the optimal recognition accuracy. To exclude this potential problem, we re-score the labels of both Swin-B and Mixup at varying levels of our classification hierarchy defined in Sec. \ref{sec:human_labeling}, with the full class hierarchy of IRUO in Fig. \ref{fig:hierarchy}. For each level, we re-score human and model predictions, combining high-level predictions into single labels as the classification level decreases, e.g., to score at level 2, we combine labels for \textit{bird}, \textit{parrot}, and \textit{poultry} into a single \textit{bird} label and re-score.The re-scored results for Swin-Mixup and human observers at each classification level are shown in Fig. \ref{fig:top_humans}. The difference between accuracy achieved by Swin-Mixup and humans is not substantially different as we vary the hierarchy level, suggesting that our human population was not significantly disadvantaged when we use the highest level of the hierarchy (i.e., the most granular class labels).

\subsection{Further analysis: are models robust to diffuse occlusion?}\label{sec:diffuse}
\begin{figure}[h!]
\begin{center}
\begin{tabular}{ccc}
\includegraphics[width=0.27\linewidth]{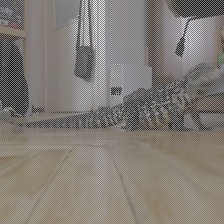}&
\includegraphics[width=0.27\linewidth]{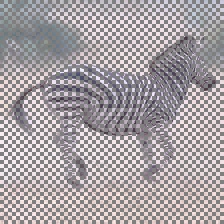}&
\includegraphics[width=0.27\linewidth]{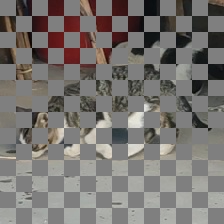}\\
(a)&(b)&(c)
\end{tabular}
\caption{Images showing levels of diffuse occlusion in the IRUO-Diffuse subset, i.e., occlusion comprised of small occlusions visually dividing objects, applied, where occlusion patterns contain (a) 1x1 pixel occluders, (b) 4x4-pixel occluders, (c) 16x16-pixel occluders}
\label{fig:diffuse_samples2}
\end{center}
\end{figure}

The results in Sec. \ref{sec:human_labeling} suggest that contemporary DNNs (e.g., Swin) achieve nearly the same accuracy under occlusion as humans. In this section we investigate whether DNN robustness depends upon the \textit{properties} of the occluding objects. Our investigation reveals that DNN occlusion-robustness depends strongly upon at least one property, which we term "diffuseness". We define diffuseness as the average proportion of neighbors of occluding pixels that are \textbf{not} occluders. Fig. \ref{fig:diffuse_samples2} presents three images with synthetic occlusions, and where the occlusions have varying levels of diffuseness (e.g., (a) is most, while (b) is least). Our primary hypothesis is that the occlusion-robustness of DNNs reduces as diffuseness increases, suggesting that there is still substantial potential to improve DNNs for some types of occlusion.

\textbf{Experimental Design.} We conduct two experiments to evaluate the effect of diffuse occluders on the occlusion-robustness of DNNs. To conduct these experiments, we leverage our findings in Sec. \ref{sec:synth} where we found that synthetic occlusions are a reasonable substitute for real-world occlusions if we wish to rank-order the performance of models under occlusion. Using this finding, we created an additional partition of the IRUO dataset, termed \textit{IRUO-Diffuse} (see Sec. \ref{sec:dataset}), wherein we generate several types of synthetic occlusions with varying types and levels of "diffuseness" to evaluate both experiments.

\begin{figure}[h!]
\begin{center}
\begin{tabular}{ccc}
\includegraphics[width=0.27\linewidth]{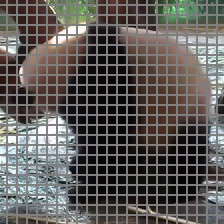}&
\includegraphics[width=0.27\linewidth]{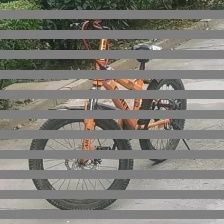}&
\includegraphics[width=0.27\linewidth]{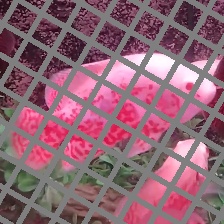}\\
(a)&(b)&(c)
\end{tabular}
\caption{Images showing types of diffuse occlusion included in the IRUO-Diffuse subset. Occlusion types: (a) cross-hatch (grid) occlusions, (b) horizontal line occlusions, and (c) occlusions by lines at non-right angles}
\label{fig:diffuse_samples}
\end{center}
\end{figure}

Our first experiment evaluates whether DNNs might be less robust to diffuse occlusion than solid occluders. To conduct this experiment, in IRUO-Diffuse, we created solid gray box occluders and two types of diffuse gray occluders: right-angle and oblique line occluders, both of which are in Fig. \ref{fig:diffuse_samples}. Solid box occluders are randomly generated with a random size and center point location, while the diffuse occluders are randomly generated with randomly selected line width and spacing, along with randomly selected angles for oblique occluders. Full details of image generation in IRUO-Diffuse are in the Appendix. By varying these parameters, we generate both solid and diffuse occlusions at each occlusion level, allowing us to compare the effect of diffuse occluders on the occlusion robustness of DNNs. Since we are able to control the precise level of occlusion generated via this method, we can also evaluate model accuracy for more granular levels of occlusion than in previous sections. In Sections \ref{sec:results}, \ref{sec:synth}, and \ref{sec:human_study} we utilized three occlusion levels, whereas here we utilize five occlusion levels: 0\% (unoccluded imagery), 10-30\%, 30-50\%, 50-70\%, and 70-90\% occlusion.

Our second experiment evaluates the effect of the \textit{level} of diffuseness on the occlusion-robustness of DNNs, since the first experiment provides evidence that DNNs are indeed less robust to diffuse occlusion. We include in IRUO-Diffuse images containing occluders with each of five specific levels of diffuseness: denoted 1, 2, 4, 8, or 16. Each number corresponds to the size (in pixels) of evenly-spaced contiguous groups of occluding pixels, so that \textit{larger numbers correspond to lower diffuseness}. Examples of images illustrating these occluder patterns are presented in Fig. \ref{fig:diffuse_samples2}. Like in the first experiment, the use of synthetic occlusions allows us to generate more granular levels of occlusion than in previous sections; in this experiment, we utilize four levels of occlusion: 0\% (unoccluded imagery), 25\%, 50\%, and 75\% occlusion. Additional details of the generation of this subset of IRUO-Diffuse are in the Appendix.

For both experiments, to evaluate the \textit{robustness} -- as opposed to pure accuracy -- of Swin, we needed to compare Swin to humans. We recruited additional human observers (distinct from Sec. \ref{sec:human_study}) to label two subsets of the IRUO-diffuse subset: one for each of the two experiments outlined above. For \textit{each experiment} we recruited five human observers, and each observer rated 50 images, for a total of 250 rated images per experiment. Human responses are collected using the same method described in Sec. \ref{sec:human_design}. Similar to the generation of IRUO-HTS, we subsample 2-3 images from each class at each occlusion level for each type of occlusion, resulting in a total of approximately 250 images for each of two sets: one containing images with the solid and diffuse occluders described in the first experiment above and in Fig. \ref{fig:diffuse_samples}, and the other containing images with the various-sized occluders described in the second experiment above and in Fig. \ref{fig:diffuse_samples2}. For both experiments, the reported accuracies for Swin and human observers at occlusion level 0 are the same accuracies as in Sec. \ref{sec:human_study}, as the images with occlusion level 0 (0\% occlusion) are unchanged in this section.

\begin{figure}[b]
\begin{center}
\includegraphics[width=0.95\linewidth]{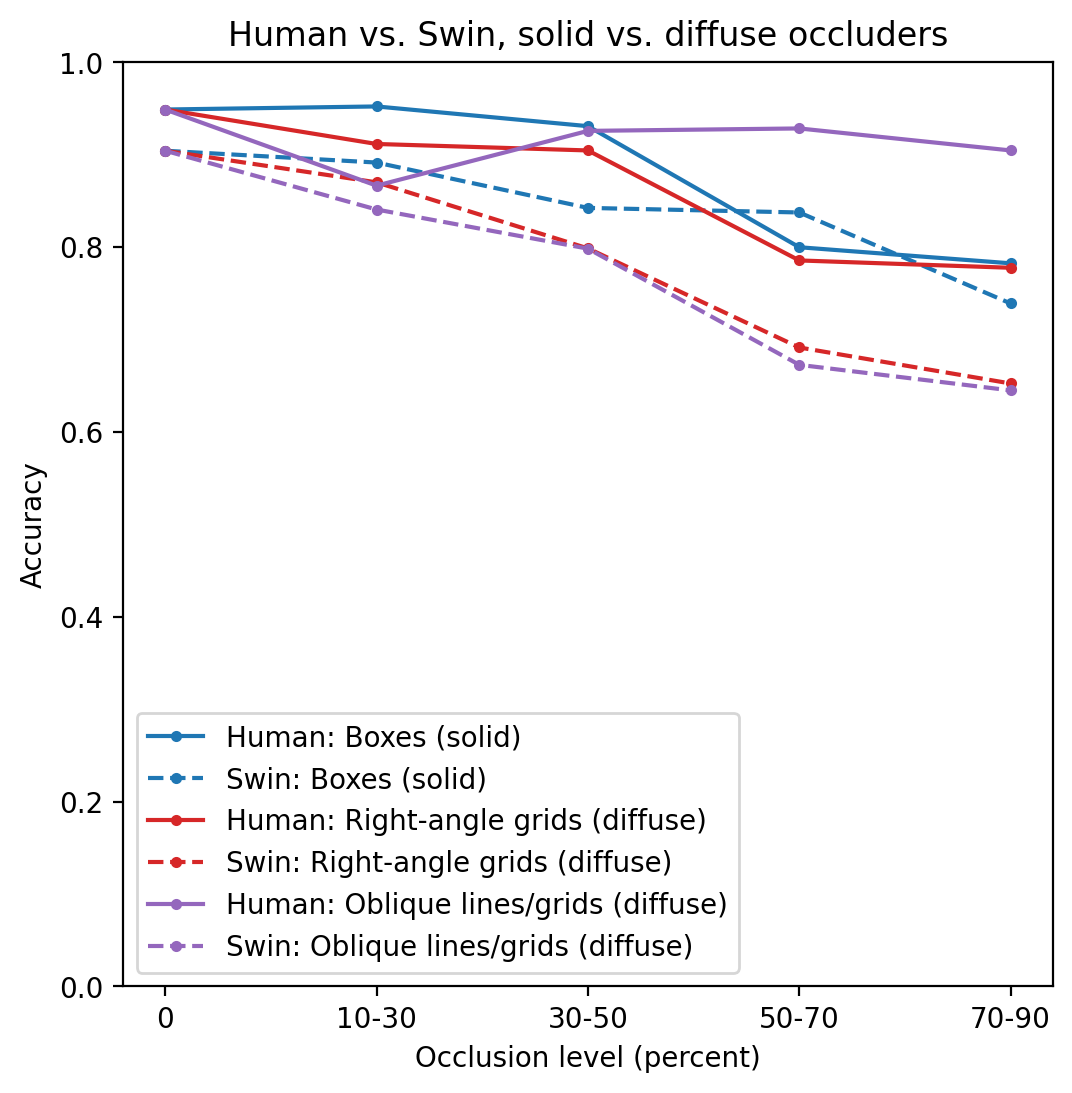}
\end{center}
\caption{Plots of results on diffuse occlusion, including right-angle and oblique grids, compared to solid gray occluders, for human observers and Swin (the best-performing model from Sec. \ref{sec:human_study})}
\label{fig:diffuse_results}
\end{figure}

\textbf{Results.} Fig. \ref{fig:diffuse_results} presents the results of our first experiment, investigating DNN occlusion-robustness for different types of occlusion by comparing Swin and human accuracy for several diffuse occluders. The gap between human observers and Swin is minimal for solid box occluders at higher levels of occlusion and much larger for both types of diffuse occlusion (right-angle grids and oblique lines/grids, as tested). This indicates that diffuse occluders reduce the performance of models more than equivalent solid occluders at the same levels of occlusion. Further, this gap in accuracy (about 9 percent between box occlusions and the two diffuse occluders at the highest level of occlusion) is larger than the gap in accuracy between the average human observer and Swin in Sec. \ref{sec:human_study}, suggesting that diffuse occlusion contributes to the overall gap in human and DNN performance on occluded imagery.

\begin{figure}[ht]
\begin{center}
\includegraphics[width=0.95\linewidth]{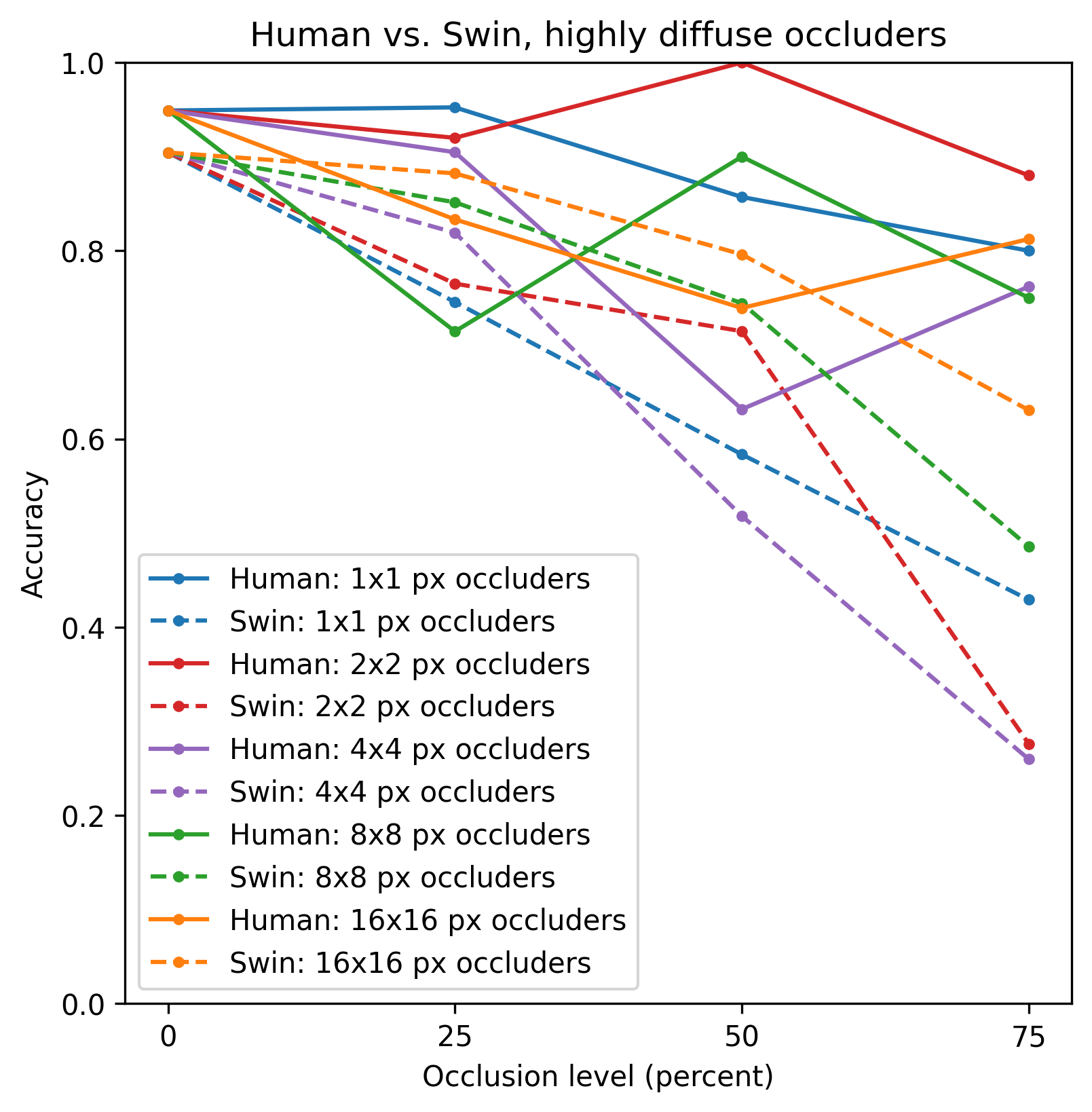}
\end{center}
\caption{Plots of results on various levels of diffuse occlusion, including pixel occlusions of 1x1- to 16x16-pixel size, for human observers and Swin (the best-performing model from Sec. \ref{sec:human_study})}
\label{fig:diffuse_results2}
\end{figure}

 Fig. \ref{fig:diffuse_results2} presents the results of our second experiment, where we observe that smaller occluders (those with pixel sizes of 4 or less) degrade the accuracy of Swin, as compared to humans, more than larger occluders (those with pixel size of 8 or more). These larger gaps in accuracy between Swin and humans for smaller (i.e., more diffuse) occluders indicate that the level of diffuseness of occluders has an inverse relationship with model accuracy at higher levels of occlusion. In Fig. \ref{fig:diffuse_results2}, the results corresponding to 16x16-pixel occlusions (the least diffuse in this set) suggest minimal difference between human and Swin accuracy, on par with that of solid occluders (as in Fig. \ref{fig:diffuse_results}). Some of the difference in accuracy between human observers and Swin, especially on the diffuse occlusions in Fig. \ref{fig:diffuse_samples2} may result from the choice of patch sizes in Swin (typically 4x4 pixels each); larger occluders are more likely to coincide with entire patches than smaller occluders, possibly allowing models to ignore these occlusions more effectively.

\begin{figure*}[ht]
 \begin{center}
\begin{tabular}{ccccc}
(16)&
\includegraphics[width=0.2\linewidth]{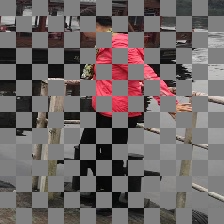}&
\includegraphics[width=0.2\linewidth]{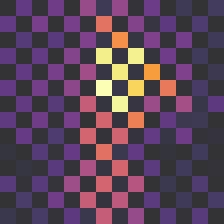}&
\includegraphics[width=0.2\linewidth]{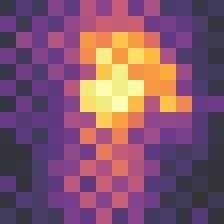}&
\includegraphics[width=0.2\linewidth]{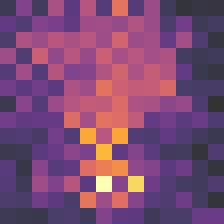}
\\
(8)&
\includegraphics[width=0.2\linewidth]{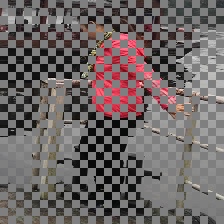}&
\includegraphics[width=0.2\linewidth]{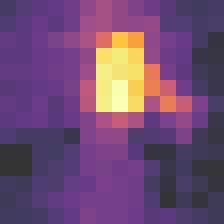}&
\includegraphics[width=0.2\linewidth]{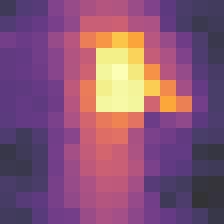}&
\includegraphics[width=0.2\linewidth]{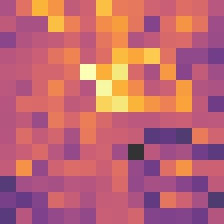}
\\
(4)&
\includegraphics[width=0.2\linewidth]{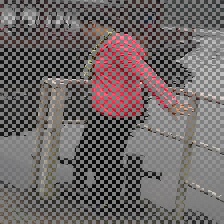}&
\includegraphics[width=0.2\linewidth]{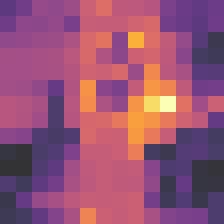}&
\includegraphics[width=0.2\linewidth]{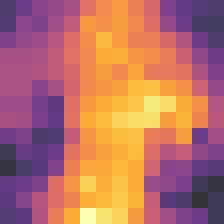}&
\includegraphics[width=0.2\linewidth]{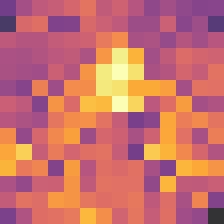}
\\
(size, px)&(a)&(b)&(c)&(d)\\
\end{tabular}
 \end{center}
\caption{Sample images of a person (column (a)) at 50\% occlusion, with various occluder sizes (in pixels, shown in row names) with corresponding self-attention maps returned by the ViT-B model trained on IRUO at the (column (b)) third, (column (c)) seventh, and (column (d)) eleventh transformer stages. Self-attention maps are summed over all points on target and all heads for each location in each layer displayed. Lighter colors (e.g., yellow) denote higher values for attention}
\label{fig:vit_attention_maps_diffuse}
\end{figure*}

To investigate why smaller occluders cause model accuracy to degrade more than larger occluders, we find the attention maps returned by Vision Transformer, a model similar to Swin in its attention computation but with more interpretable image-wide attention output. In Fig. \ref{fig:vit_attention_maps_diffuse}, we plot attention maps at stages 3, 7, and 11 (identical to those used in the procedure used for Fig. \ref{fig:vit_attention_maps}) for an identical image with occluders of size 4, 8, and 16 pixels. The results here are representative of those for most images; additional images and their corresponding attention maps are located in the Appendix. We observe that images show high values of attention to points that are not occluded when occluders are large, and that this attention becomes much lower as the size of occluders decreases, particularly as occluders become smaller than the patch size of the Vision Transformer. We hypothesize that when occluders become smaller than the patch size of models (assuming the same level of occlusion and therefore higher number of occluders), these occluders contaminate patch-level computations in transformer models and greatly reduce their ability to make accurate predictions.

\section{Conclusion}
Consistent with existing literature, we observe that occlusion is a significant challenge for deep learning models that are otherwise able to achieve human accuracy on visual recognition tasks. We observe that some models designed to perform well under conditions of partial occlusion may rely on assumptions, including a small number of classes and known occlusion types, that do not necessarily reflect real-world applications. The benchmark results presented in Sec. \ref{sec:benchmark} demonstrate that, while some models, such as Swin (\cite{liu_swin_2021}) nearly achieve human accuracy on occluded data, some types of occlusion, including diffuse occlusion, are still challenging for these models.

In this work, we provide evidence for the following:
\begin{enumerate}
    \item \textbf{Comparison of occlusion-robust models.}
    Transformer models outperform convolutional models, typically by a large margin, on both occluded and unoccluded imagery.
    Swin with Mixup augmentation achieved the best results across all levels of occlusion, including none.
    Augmentations suggested to work well for occlusion (e.g., Mixup, CutMix, Deep Feature Augmentation) provide mixed benefit.
    Models especially designed for occlusion (e.g., CompositionalNet) perform worse than all other models tested on our benchmark.
    
    \item \textbf{Comparison of real and synthetic imagery for occlusion evaluation.}
    Synthetic occlusions are a reasonable substitute for real occlusions when comparing models on occluded imagery. As shown in Sec. \ref{sec:synth}, we find that the rank-order of models is highly similar, regardless of the type of occlusion used. This finding reduces the difficulty of data collection to evaluate model performance on various levels of occlusion.
    
    \item \textbf{Robustness of DNN models.}
    We find that humans perform somewhat better than the top-performing model, Swin with Mixup; however, this gap in performance is somewhat small. This result suggests that some improvement can be made to this model, but the overall potential for improvement is likely limited.
    Model robustness depends on the properties of occluders. Specifically, diffuse occluders (such as those in Fig. \ref{fig:diffuse_samples}) are more challenging for models, even state-of-the-art models such as Swin, than solid occluders at equivalent levels of occlusion.

\end{enumerate}

This benchmark does not stratify results by certain \textit{types} of real occlusion, including inter- and intra-class occlusion, because OVIS (and, by extension, IRUO) does not contain these distinctions, and additional manual labeling is beyond the scope of this publication. Notably, intra-class occlusion (e.g., in Fig. \ref{fig:blur_samples}(a)) poses a difficult challenge in evaluation on occlusion; an object in the background may be mostly occluded by an object of the same class that is mostly visible to the camera. More recent datasets, including HOOT (\cite{sahin_hoot_2023}), include labels making these distinctions, but they do not serve as full replacements for IRUO, since they contain more limited stratifications of occlusion levels. The authors propose additional experiments such as those evaluating models on specific types of occlusions as future work.

\section*{Acknowledgment}
This research was supported by the U.S. Army Combat Capabilities Development Command (DEVCOM), C5ISR Center, RTI Directorate via Grants W15P7T-19-D-0082 and W909MY-19-F-0095. The authors would like to thank the participants in our human research studies, who will remain anonymous.

\section*{Data Availability}
Code for data generation, tested models, and other results, are available online at https://github.com/kalebkassaw/iruo. Usage of this repository requires the original OVIS dataset (\cite{qi_occluded_2022}) to be downloaded.

\begin{appendices}

\begin{landscape}
\section{Full result tables for model evaluation on real and synthetic occlusion}\label{secA1}
\begin{table}[h!]
    \resizebox{\linewidth}{!}{%
    \begin{tabular}{cc|c|cccccc|cccccc}
         \hline
         \multirow{3}{*}{Model} & \multirow{3}{*}{Augmentation} & \multicolumn{13}{c}{Accuracy} \\ 
         &&0 (None)&\multicolumn{6}{c}{1 (0-50\%)} &\multicolumn{6}{c}{2 (50-100\%)}\\
         &&-&Real&White&Black&Noise&Texture&Objects&Real&White&Black&Noise&Texture&Objects\\
         \hline
         VGG-16 (\cite{simonyan_very_2015} & -  & 0.755&0.609&0.706&0.707&0.688&0.668&0.593&0.430&0.493&0.518&0.474&0.401&0.480\\
         CompositionalNet (\cite{kortylewski_compositional_2020}  & - & 0.667&0.508&0.635&0.646&0.625&0.600&0.565&0.321&0.436&0.464&0.437&0.249&0.437\\
         ViT (\cite{dosovitskiy_image_2021} & - & 0.800&0.659&0.411&0.367&0.429&0.363&0.376&0.477&0.317&0.258&0.366&0.156&0.354\\
         ResNet-50 (\cite{he_deep_2016} & - & 0.793&0.651&0.744&0.755&0.734&0.701&0.587&0.468&0.503&0.540&0.460&0.405&0.413\\
         \hline
         \multirow{4}{*}{ResNeXt-50 (\cite{xie_aggregated_2017})} & - & 0.811&0.665&0.769&0.777&0.742&0.657&0.666&0.496&0.547&0.574&0.514&0.267&0.510\\
         & Mixup (\cite{zhang_mixup_2018}) & 0.793&0.655&0.761&0.756&0.732&0.670&0.669&0.490&0.538&0.548&0.528&0.357&0.523\\
         & CutMix (\cite{yun_cutmix_2019}) & 0.785&0.640&0.753&0.755&0.733&0.705&0.688&0.480&0.527&0.556&0.550&0.484&0.545\\
         & Deep Feature (\cite{cen_deep_2021})& 0.781&0.643&0.737&0.736&0.713&0.552&0.638&0.479&0.519&0.519&0.489&0.188&0.482\\
         \hline
         \multirow{3}{*}{DeiT (\cite{touvron_training_2021})} & - & 0.862&0.729&0.854&0.862&0.858&0.708&0.842&0.574&0.728&0.748&0.785&0.359&0.731\\
         & Mixup (\cite{zhang_mixup_2018}) & 0.868&0.728&0.863&0.868&0.859&0.733&0.851&0.568&0.730&0.766&0.799&0.411&0.736\\
         & CutMix (\cite{yun_cutmix_2019}) & 0.863&0.729&0.861&0.862&0.859&0.680&0.848&0.567&0.730&0.766&0.784&0.357&0.740\\
         \hline
         \multirow{3}{*}{Swin (\cite{liu_swin_2021})} & - & 0.891&0.751&0.883&0.882&0.882&0.870&0.855&0.594&0.722&0.785&0.771&0.762&0.698\\
         & Mixup (\cite{zhang_mixup_2018}) & 0.894&0.760&0.882&0.882&0.878&0.873&0.863&0.597&0.758&0.789&0.793&0.754&0.731\\
         & CutMix (\cite{yun_cutmix_2019}) & 0.885&0.744&0.872&0.863&0.869&0.864&0.860&0.581&0.743&0.780&0.772&0.731&0.726\\
         \hline
    \end{tabular}
    }
    \caption{Table showing raw accuracy scores for each model configuration tested with each type of occlusion. For each level of occlusion greater than 0, we stratified occlusion into one of many types: real occlusion plus five types of artificial occlusions commonly used in the literature}
    \label{tab:full_accuracy}
\end{table}

\begin{table}[h!]
    \resizebox{\linewidth}{!}{%
    \begin{tabular}{cc|c|cccccc|cccccc}
         \hline
         \multirow{3}{*}{Model} & \multirow{3}{*}{Augmentation} & \multicolumn{13}{c}{Accuracy} \\ 
         &&0 (None)& \multicolumn{6}{c}{1 (0-50\%)} &\multicolumn{6}{c}{2 (50-100\%)}\\
         &&-&Real&White&Black&Noise&Texture&Objects&Real&White&Black&Noise&Texture&Objects\\
         \hline
         VGG-16 (\cite{simonyan_very_2015}) & -  & 13&13&12&12&12&10&11&13&12&12&11&7&11\\
         CompositionalNet (\cite{kortylewski_compositional_2020})  & - & 14&14&13&13&13&12&13&14&13&13&13&12&12\\
         ViT (\cite{dosovitskiy_image_2021}) & - & 8&8&14&14&14&14&14&11&14&14&14&14&14\\
         ResNet (\cite{he_deep_2016}) & - & 9.5&10&10&9.5&8&7&12&12&11&10&12&6&13\\
         \hline
         \multirow{4}{*}{ResNeXt (\cite{xie_aggregated_2017})} & - & 7&7&7&7&7&11&9&7&7&7&9&11&9\\
         & Mixup (\cite{zhang_mixup_2018}) & 9.5&9&8&8&10&9&8&8&8&9&8&9.5&8\\
         & CutMix (\cite{yun_cutmix_2019}) & 11&12&9&9.5&9&6&7&9&9&8&7&4&7\\
         & Deep Feature (\cite{cen_deep_2021})& 12&11&11&11&11&13&10&10&10&11&10&13&10\\
         \hline
         \multirow{3}{*}{DeiT (\cite{touvron_training_2021})} & - & 6&4.5&6&5.5&6&5&6&4&5&6&3&8&3.5\\
         & Mixup (\cite{zhang_mixup_2018}) & 4&6&4&3&4.5&4&4&5&3.5&4.5&1&5&2\\
         & CutMix (\cite{yun_cutmix_2019}) & 5&4.5&5&5.5&4.5&8&5&6&3.5&4.5&4&9.5&1\\
         \hline
         \multirow{3}{*}{Swin (\cite{liu_swin_2021})} & - & 2&2&1&1.5&1&2&3&2&6&2&6&1&6\\
         & Mixup (\cite{zhang_mixup_2018}) & 1&1&2&1.5&2&1&1&1&1&1&2&2&3.5\\
         & CutMix (\cite{yun_cutmix_2019}) & 3&3&3&4&3&3&2&3&2&3&5&3&5\\
         \hline
         \multicolumn{2}{c}{Rank ordering $Q$ (\cite{friedman_use_1937})} &-& \multicolumn{6}{c}{$Q=71.0$; $p=5.0\times10^{-10}$} & \multicolumn{6}{c}{$Q=65.5$; $p=5.3\times10^{-9}$}\\
         \hline
    \end{tabular}}
    \caption{Table showing the rankings of models per each experimental setup with corresponding Friedman $Q$ values and their statistical significance. The type of occlusion does not appear to affect rankings of models in a meaningful way, as determined by the high values of $Q$ for both occlusion levels 1 and 2}
    \label{tab:full_rank}
\end{table}

\clearpage

\end{landscape}

\section{Details of training}
Each model is pretrained on ImageNet (\cite{deng_imagenet_2009}) and fine-tuned on the subset of the IRUO dataset containing no occlusion. (This setup is consistent with other training schemes, e.g., in \cite{kortylewski_compositional_2020, cen_deep_2021}.) For each model, we use a stochastic gradient descent optimizer, and we train until convergence to highest validation accuracy, up to 50 training epochs. Most models and augmentations are derived from MMClassification, a PyTorch-based deep learning package. We use a grid search to find ideal learning rates, momentums, and weight decays for each model and augmentation configuration. For CompositionalNet (\cite{kortylewski_compositional_2020}), we additionally initialize the von Mises-Fisher clusters, similarity matrices, and mixture models on IRUO images at occlusion level 0, and we optimize the weights of cluster center and mixture model losses during training. For Deep Feature Augmentation (\cite{cen_deep_2021}), we use our best-performing ResNeXt-50 model trained on IRUO and fine-tune for up to 20 additional epochs, retaining highest validation accuracy. We also perform a grid search over learning rate, deep vector probability, and deep vector weight. The deep feature vectors and occluder images used in \cite{cen_deep_2021} are not publicly available; therefore, we use a different set of objects cropped from the COCO dataset \cite{lin_microsoft_2015} containing only objects that are otherwise not part of IRUO, e.g., stop signs and refrigerators.

\section{Additional details of human data collection}
\label{sec:human_appendix}
To mitigate and track errors due to selection of incorrect objects, we ask humans to label which object they are selecting by clicking a single point on the image. This choice is guided without directly revealing which object to label; a cross is placed at the center of the screen to remind participants that objects are expected to be near this point, but it is not moved from this fixed point, regardless of where objects are actually centered. 

\section{Calculation of Friedman Q-value}
\label{sec:friedman_calc}

For the model rank comparison described in \ref{sec:stats} and used in Sec. \ref{sec:synth}, we opt to use the Friedman $Q$-value in \cite{friedman_use_1937}, defined below in Eq. \ref{eq:rank}. Specifically, we compare the rank ordering of models in our evaluations using real and synthetic occlusion, as opposed to using raw scores that may not be as informative for different types of occluders. We evaluate $Q$ in the formula 
\begin{equation}
Q = \frac{12n}{k(k+1)}\sum_{j=1}^{k}(\bar{r_j}-\frac{k+1}{2})^2
\label{eq:rank}
\end{equation}
where $n$ is the number of models evaluated for each occlusion type, $k$ is the number of occlusion types, and $r_j$ is the average rank order assigned to each model over all occlusion types. We split this evaluation into two parts, determining whether models have specific rank orders at various occlusion levels, regardless of the type of occlusion. For $k>4$ or $n>15$, $Q$ has a chi-squared null distribution with $k$ degrees of freedom (\cite{friedman_use_1937}).%

\section{Details of synthetic data generation}\label{secA2}
Synthetic occlusions in IRUO-Synthetic are one of five types: 1) black boxes, 2) white boxes, 3) noise boxes generated from a uniform distribution over red, green, and blue channels, 4) zebra-print texture boxes, and 5) objects from the COCO dataset. These synthetic occlusion types are based on those used in the studies in \cite{kortylewski_compositional_2020, zhu_robustness_2019, naseer_intriguing_2021, chen_transmix_2022}.
We generate boxes by randomly selecting a box center point (from a uniform distribution over image dimensions) and an image size (from a Gaussian distribution of both dimensions, mean $S/2$, standard deviation $3S/10$, where $S$ is the side dimension of the square input image). While these parameters result in images containing object occlusion levels spread out over the 0-100 percent range, some images containing unoccluded or fully occluded objects result. Occlusions that cover less than 5 percent or greater than 95 percent of images are removed.
Synthetic objects are resized to 1/4 of image area and placed at random on top of objects, with center points randomly distributed over the middle half of each dimension.
To balance computation time with result accuracy, we generate one occluded image per unoccluded IRUO image per type of synthetic occlusion, and we calculate object occlusion percentages by calculating the intersection of occlusions and original object masks, dividing it by the object mask area, and subtracting this value from 1.

\section{Details of diffuse synthetic data generation}
Because we determine that synthetic occlusions can serve as an appropriate proxy for real-world occlusion scenes, we generate IRUO-Diffuse, another dataset comprised of synthetic occlusions to evaluate the effect of diffuse occluders, e.g., sparsely arranged occlusions such as fences and leaves. To compare the effects of solid versus diffuse occluders on human and model accuracy, we generate occlusion boxes using largely similar parameters to that in Sec. \ref{secA2}, except with size standard deviation $4S/10$, where $S$ is the side dimension of the square input image, to account for the overall lower amount of occlusion resulting from diffuse occluders in similar window sizes. If images have much higher levels of background occlusion than equivalent solid occluders (e.g., the percentage of background occlusion is more than twice that of object occlusion), they are re-generated.

To measure the effect of \textit{diffuseness}, i.e., the numerical sparsity of occluders on human and model accuracy, we generate diffuse occlusions using simple grid patterns comprising of 25, 50, and 75 percent occlusion. These grid patterns are expressed as tiles that are repeated over an entire image, and a value of 1 returns an occluded pixel, while a value of 0 returns the original image pixel. To generate data over multiple levels of diffuseness, we upscale these tiles (such that fewer, larger tiles are used) by 2x in both spatial dimensions.

Sample occluder tile masks are as follows, denoted $O_{x,y}$, where $x$ denotes the percentage of object occlusion and $y$ denotes the number of 2x upscalings:

\begin{table}[h!]
    \resizebox{\linewidth}{!}{%
    \begin{tabular}{ccc}
         $O_{25, 0} = \begin{bmatrix}
             1 & 0 \\ 0 & 0
         \end{bmatrix}$ & 
         $O_{50, 0} = \begin{bmatrix}
             1 & 0 \\ 0 & 1
         \end{bmatrix}$ & 
         $O_{75, 0} = \begin{bmatrix}
             1 & 1 \\ 0 & 1
         \end{bmatrix}$
    \end{tabular}}
\end{table}

\begin{table}[h!]
    \resizebox{\linewidth}{!}{%
    \begin{tabular}{cc}
         $O_{50, 1} = \begin{bmatrix}
             1 & 1 & 0 & 0 \\ 1 & 1 & 0 & 0 \\ 0 & 0 & 1 & 1 \\ 0 & 0 & 1 & 1
         \end{bmatrix}$ & 
         $O_{50, 2} = \begin{bmatrix}
             1 & 1 & 1 & 1 & 0 & 0 & 0 & 0 \\ 1 & 1 & 1 & 1 & 0 & 0 & 0 & 0 \\ 1 & 1 & 1 & 1 & 0 & 0 & 0 & 0 \\ 1 & 1 & 1 & 1 & 0 & 0 & 0 & 0 \\ 0 & 0 & 0 & 0 & 1 & 1 & 1 & 1 \\ 0 & 0 & 0 & 0 & 1 & 1 & 1 & 1 \\ 0 & 0 & 0 & 0 & 1 & 1 & 1 & 1 \\ 0 & 0 & 0 & 0 & 1 & 1 & 1 & 1
         \end{bmatrix}$
    \end{tabular}}
\end{table}

\section{Additional Vision Transformer attention maps}
\label{sec:supplemental_maps}

In this section we provide additional images such as those in Fig. \ref{fig:vit_attention_maps_diffuse} to corroborate results in this section. As shown in these figures, there exists a pattern in which Vision Transformer models are able to attend to pixels on target and successfully ignore occluders when they are large, and that this relative attention becomes less clear as the occluders get smaller.

\begin{figure}[ht]
 \begin{center}
\begin{tabular}{ccccc}
(16)&
\includegraphics[width=0.14\linewidth]{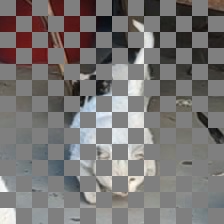}&
\includegraphics[width=0.14\linewidth]{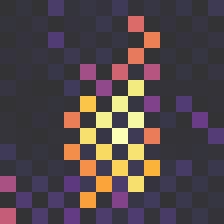}&
\includegraphics[width=0.14\linewidth]{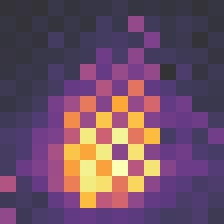}&
\includegraphics[width=0.14\linewidth]{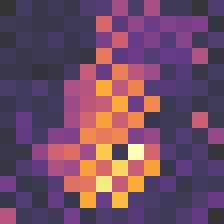}
\\
(8)&
\includegraphics[width=0.14\linewidth]{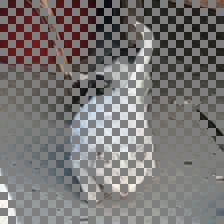}&
\includegraphics[width=0.14\linewidth]{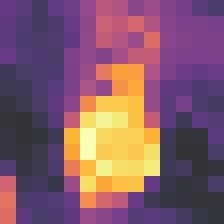}&
\includegraphics[width=0.14\linewidth]{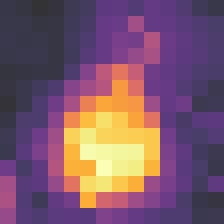}&
\includegraphics[width=0.14\linewidth]{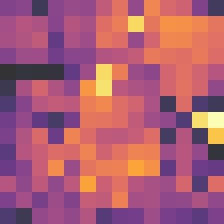}
\\
(4)&
\includegraphics[width=0.14\linewidth]{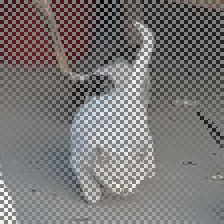}&
\includegraphics[width=0.14\linewidth]{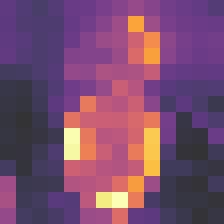}&
\includegraphics[width=0.14\linewidth]{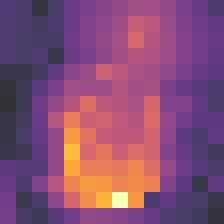}&
\includegraphics[width=0.14\linewidth]{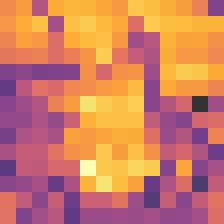}
\\
(size, px)&(a)&(b)&(c)&(d)\\
\end{tabular}
 \end{center}
\caption{Sample images of a cat (column (a)) at 50\% occlusion, with various occluder sizes (in pixels, shown in row names) with corresponding self-attention maps returned by the ViT-B model trained on IRUO at the (column (b)) third, (column (c)) seventh, and (column (d)) eleventh transformer stages. Self-attention maps are summed over all points on target and all heads for each location in each layer displayed. Higher values (e.g., yellow) denote higher values for attention}
\label{fig:vit_attention_maps_diffuse_a1}
\end{figure}

\begin{figure}[ht]
 \begin{center}
\begin{tabular}{ccccc}
(16)&
\includegraphics[width=0.14\linewidth]{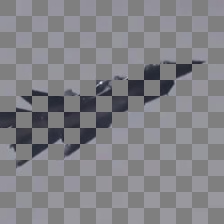}&
\includegraphics[width=0.14\linewidth]{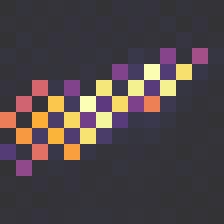}&
\includegraphics[width=0.14\linewidth]{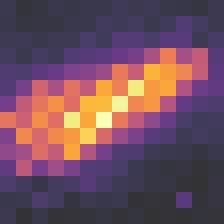}&
\includegraphics[width=0.14\linewidth]{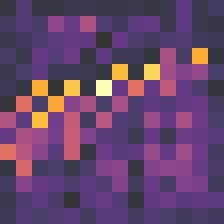}
\\
(8)&
\includegraphics[width=0.14\linewidth]{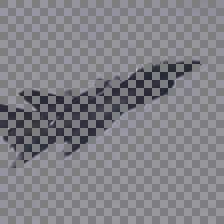}&
\includegraphics[width=0.14\linewidth]{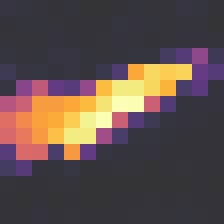}&
\includegraphics[width=0.14\linewidth]{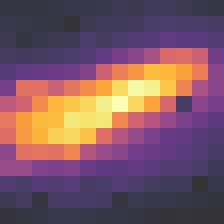}&
\includegraphics[width=0.14\linewidth]{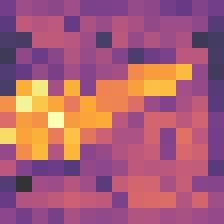}
\\
(4)&
\includegraphics[width=0.14\linewidth]{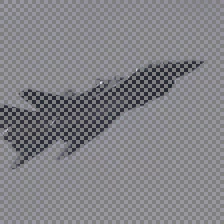}&
\includegraphics[width=0.14\linewidth]{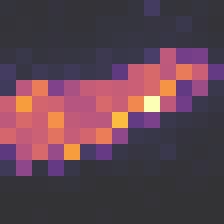}&
\includegraphics[width=0.14\linewidth]{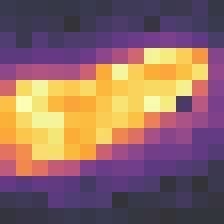}&
\includegraphics[width=0.14\linewidth]{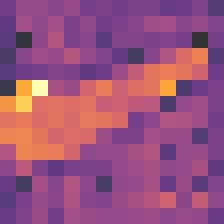}
\\
(size, px)&(a)&(b)&(c)&(d)\\
\end{tabular}
 \end{center}
\caption{Sample images of an airplane (column (a)) at 50\% occlusion, with various occluder sizes (in pixels, shown in row names) with corresponding self-attention maps returned by the ViT-B model trained on IRUO at the (column (b)) third, (column (c)) seventh, and (column (d)) eleventh transformer stages. Self-attention maps are summed over all points on target and all heads for each location in each layer displayed. Higher values (e.g., yellow) denote higher values for attention}
\label{fig:vit_attention_maps_diffuse_a2}
\end{figure}

\begin{figure}[ht]
 \begin{center}
\begin{tabular}{ccccc}
(16)&
\includegraphics[width=0.14\linewidth]{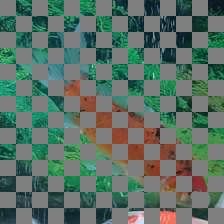}&
\includegraphics[width=0.14\linewidth]{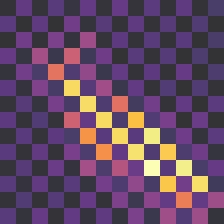}&
\includegraphics[width=0.14\linewidth]{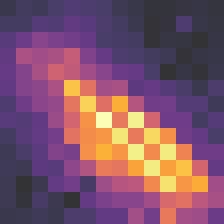}&
\includegraphics[width=0.14\linewidth]{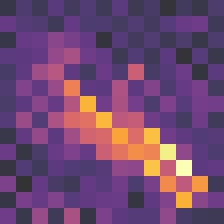}
\\
(8)&
\includegraphics[width=0.14\linewidth]{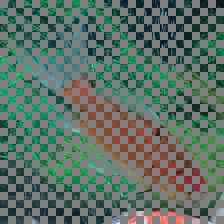}&
\includegraphics[width=0.14\linewidth]{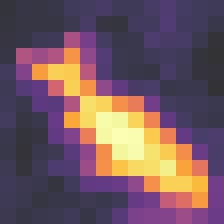}&
\includegraphics[width=0.14\linewidth]{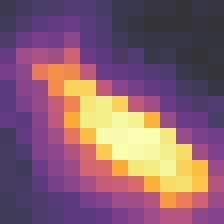}&
\includegraphics[width=0.14\linewidth]{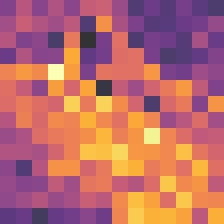}
\\
(4)&
\includegraphics[width=0.14\linewidth]{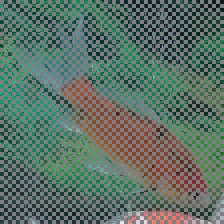}&
\includegraphics[width=0.14\linewidth]{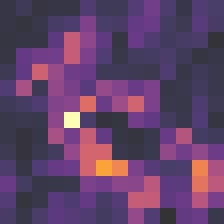}&
\includegraphics[width=0.14\linewidth]{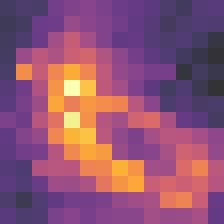}&
\includegraphics[width=0.14\linewidth]{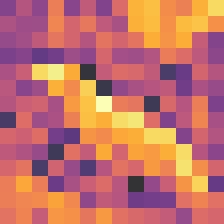}
\\
(size, px)&(a)&(b)&(c)&(d)\\
\end{tabular}
 \end{center}
\caption{Sample images of a fish (column (a)) at 50\% occlusion, with various occluder sizes (in pixels, shown in row names) with corresponding self-attention maps returned by the ViT-B model trained on IRUO at the (column (b)) third, (column (c)) seventh, and (column (d)) eleventh transformer stages. Self-attention maps are summed over all points on target and all heads for each location in each layer displayed. Higher values (e.g., yellow) denote higher values for attention}
\label{fig:vit_attention_maps_diffuse_a3}
\end{figure}

\begin{figure}[ht]
 \begin{center}
\begin{tabular}{ccccc}
(16)&
\includegraphics[width=0.14\linewidth]{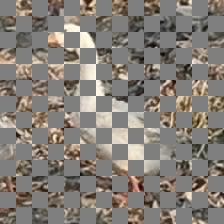}&
\includegraphics[width=0.14\linewidth]{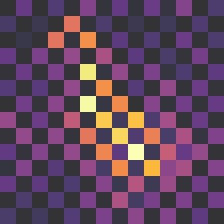}&
\includegraphics[width=0.14\linewidth]{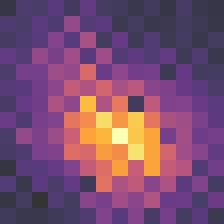}&
\includegraphics[width=0.14\linewidth]{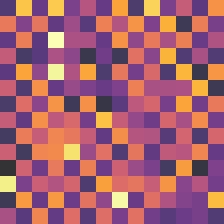}
\\
(8)&
\includegraphics[width=0.14\linewidth]{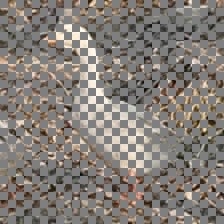}&
\includegraphics[width=0.14\linewidth]{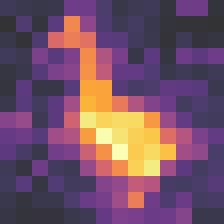}&
\includegraphics[width=0.14\linewidth]{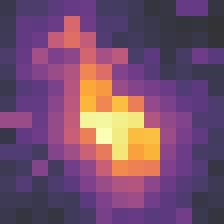}&
\includegraphics[width=0.14\linewidth]{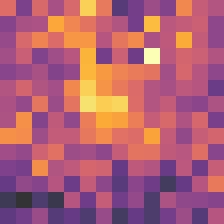}
\\
(4)&
\includegraphics[width=0.14\linewidth]{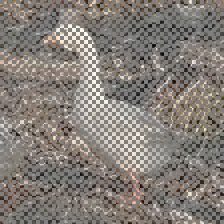}&
\includegraphics[width=0.14\linewidth]{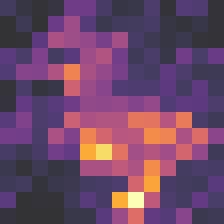}&
\includegraphics[width=0.14\linewidth]{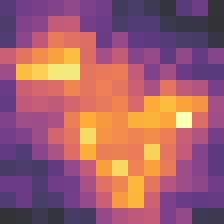}&
\includegraphics[width=0.14\linewidth]{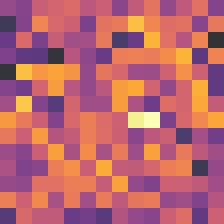}
\\
(size, px)&(a)&(b)&(c)&(d)\\
\end{tabular}
 \end{center}
\caption{Sample images of a person (column (a)) at 50\% occlusion, with various occluder sizes (in pixels, shown in row names) with corresponding self-attention maps returned by the ViT-B model trained on IRUO at the (column (b)) third, (column (c)) seventh, and (column (d)) eleventh transformer stages. Self-attention maps are summed over all points on target and all heads for each location in each layer displayed. Higher values (e.g., yellow) denote higher values for attention}
\label{fig:vit_attention_maps_diffuse_a4}
\end{figure}

\end{appendices}

\clearpage
\bibliography{references}

\begin{thebibliography}{30}
\providecommand{\natexlab}[1]{#1}
\providecommand{\url}[1]{{#1}}
\providecommand{\urlprefix}{URL }
\providecommand{\doi}[1]{\url{https://doi.org/#1}}
\providecommand{\eprint}[2][]{\url{#2}}
 \bibcommenthead

\bibitem[{Bansal et~al.(2016)Bansal, Raj, and Choudhury}]{bansal_blur_2016}
Bansal R, Raj G, Choudhury T (2016) Blur image detection using {Laplacian} operator and {Open}-{CV}. In: 2016 {International} {Conference} {System} {Modeling} \& {Advancement} in {Research} {Trends} ({SMART}), pp 63--67, \doi{10.1109/SYSMART.2016.7894491}

\bibitem[{Cen et~al.(2021)Cen, Zhao, Li, and Wang}]{cen_deep_2021}
Cen F, Zhao X, Li W, et~al. (2021) Deep feature augmentation for occluded image classification. Pattern Recognition 111:107737. \doi{10.1016/j.patcog.2020.107737}, \urlprefix\url{https://www.sciencedirect.com/science/article/pii/S0031320320305409}

\bibitem[{Chen et~al.(2022)Chen, Sun, He, Torr, Yuille, and Bai}]{chen_transmix_2022}
Chen JN, Sun S, He J, et~al. (2022) {TransMix}: {Attend} to {Mix} for {Vision} {Transformers}. In: 2022 {IEEE}/{CVF} {Conference} on {Computer} {Vision} and {Pattern} {Recognition} ({CVPR}). IEEE, New Orleans, LA, USA, pp 12125--12134, \doi{10.1109/CVPR52688.2022.01182}, \urlprefix\url{https://ieeexplore.ieee.org/document/9879590/}

\bibitem[{Deng et~al.(2009)Deng, Dong, Socher, Li, Li, and Fei-Fei}]{deng_imagenet_2009}
Deng J, Dong W, Socher R, et~al. (2009) {ImageNet}: {A} large-scale hierarchical image database. In: 2009 {IEEE} {Conference} on {Computer} {Vision} and {Pattern} {Recognition}, pp 248--255, \doi{10.1109/CVPR.2009.5206848}, iSSN: 1063-6919

\bibitem[{Deng et~al.(2014)Deng, Russakovsky, Krause, Bernstein, Berg, and Fei-Fei}]{deng_scalable_2014}
Deng J, Russakovsky O, Krause J, et~al. (2014) Scalable multi-label annotation. In: Proceedings of the {SIGCHI} {Conference} on {Human} {Factors} in {Computing} {Systems}. ACM, Toronto Ontario Canada, pp 3099--3102, \doi{10.1145/2556288.2557011}, \urlprefix\url{https://dl.acm.org/doi/10.1145/2556288.2557011}

\bibitem[{DeVries and Taylor(2017)}]{devries_improved_2017}
DeVries T, Taylor GW (2017) Improved {Regularization} of {Convolutional} {Neural} {Networks} with {Cutout}. \doi{10.48550/arXiv.1708.04552}, \urlprefix\url{http://arxiv.org/abs/1708.04552}, arXiv:1708.04552 [cs]

\bibitem[{Dosovitskiy et~al.(2021)Dosovitskiy, Beyer, Kolesnikov, Weissenborn, Zhai, Unterthiner, Dehghani, Minderer, Heigold, Gelly, Uszkoreit, and Houlsby}]{dosovitskiy_image_2021}
Dosovitskiy A, Beyer L, Kolesnikov A, et~al. (2021) An {Image} is {Worth} 16x16 {Words}: {Transformers} for {Image} {Recognition} at {Scale}. \urlprefix\url{https://openreview.net/forum?id=YicbFdNTTy}

\bibitem[{Friedman(1937)}]{friedman_use_1937}
Friedman M (1937) The {Use} of {Ranks} to {Avoid} the {Assumption} of {Normality} {Implicit} in the {Analysis} of {Variance}. Journal of the American Statistical Association 32(200):675--701. \doi{10.2307/2279372}, \urlprefix\url{https://www.jstor.org/stable/2279372}, publisher: [American Statistical Association, Taylor \& Francis, Ltd.]

\bibitem[{He et~al.(2016)He, Zhang, Ren, and Sun}]{he_deep_2016}
He K, Zhang X, Ren S, et~al. (2016) Deep {Residual} {Learning} for {Image} {Recognition}. In: 2016 {IEEE} {Conference} on {Computer} {Vision} and {Pattern} {Recognition} ({CVPR}). IEEE, Las Vegas, NV, USA, pp 770--778, \doi{10.1109/CVPR.2016.90}, \urlprefix\url{http://ieeexplore.ieee.org/document/7780459/}

\bibitem[{Kortylewski et~al.(2020{\natexlab{a}})Kortylewski, He, Liu, and Yuille}]{kortylewski_compositional_2020}
Kortylewski A, He J, Liu Q, et~al. (2020{\natexlab{a}}) Compositional {Convolutional} {Neural} {Networks}: {A} {Deep} {Architecture} {With} {Innate} {Robustness} to {Partial} {Occlusion}. In: 2020 {IEEE}/{CVF} {Conference} on {Computer} {Vision} and {Pattern} {Recognition} ({CVPR}). IEEE, Seattle, WA, USA, pp 8937--8946, \doi{10.1109/CVPR42600.2020.00896}, \urlprefix\url{https://ieeexplore.ieee.org/document/9157227/}

\bibitem[{Kortylewski et~al.(2020{\natexlab{b}})Kortylewski, Liu, Wang, Zhang, and Yuille}]{kortylewski_combining_2020}
Kortylewski A, Liu Q, Wang H, et~al. (2020{\natexlab{b}}) Combining {Compositional} {Models} and {Deep} {Networks} {For} {Robust} {Object} {Classification} under {Occlusion}. In: 2020 {IEEE} {Winter} {Conference} on {Applications} of {Computer} {Vision} ({WACV}). IEEE, Snowmass Village, CO, USA, pp 1322--1330, \doi{10.1109/WACV45572.2020.9093560}, \urlprefix\url{https://ieeexplore.ieee.org/document/9093560/}

\bibitem[{Krizhevsky et~al.(2012)Krizhevsky, Sutskever, and Hinton}]{krizhevsky_imagenet_2012}
Krizhevsky A, Sutskever I, Hinton GE (2012) {ImageNet} {Classification} with {Deep} {Convolutional} {Neural} {Networks}. In: Advances in {Neural} {Information} {Processing} {Systems}, vol~25. Curran Associates, Inc., \urlprefix\url{https://proceedings.neurips.cc/paper_files/paper/2012/hash/c399862d3b9d6b76c8436e924a68c45b-Abstract.html}

\bibitem[{Lin et~al.(2015)Lin, Maire, Belongie, Bourdev, Girshick, Hays, Perona, Ramanan, Zitnick, and Dollár}]{lin_microsoft_2015}
Lin TY, Maire M, Belongie S, et~al. (2015) Microsoft {COCO}: {Common} {Objects} in {Context}. \doi{10.48550/arXiv.1405.0312}, \urlprefix\url{http://arxiv.org/abs/1405.0312}, arXiv:1405.0312 [cs]

\bibitem[{Liu et~al.(2021)Liu, Lin, Cao, Hu, Wei, Zhang, Lin, and Guo}]{liu_swin_2021}
Liu Z, Lin Y, Cao Y, et~al. (2021) Swin {Transformer}: {Hierarchical} {Vision} {Transformer} using {Shifted} {Windows}. In: 2021 {IEEE}/{CVF} {International} {Conference} on {Computer} {Vision} ({ICCV}). IEEE, Montreal, QC, Canada, pp 9992--10002, \doi{10.1109/ICCV48922.2021.00986}, \urlprefix\url{https://ieeexplore.ieee.org/document/9710580/}

\bibitem[{Miller(1995)}]{miller_wordnet_1995}
Miller GA (1995) {WordNet}: a lexical database for {English}. Communications of the ACM 38(11):39--41. \doi{10.1145/219717.219748}, \urlprefix\url{https://dl.acm.org/doi/10.1145/219717.219748}

\bibitem[{Naseer et~al.(2021)Naseer, Ranasinghe, Khan, Hayat, Shahbaz~Khan, and Yang}]{naseer_intriguing_2021}
Naseer MM, Ranasinghe K, Khan SH, et~al. (2021) Intriguing {Properties} of {Vision} {Transformers}. In: Advances in {Neural} {Information} {Processing} {Systems}, vol~34. Curran Associates, Inc., pp 23296--23308, \urlprefix\url{https://proceedings.neurips.cc/paper/2021/hash/c404a5adbf90e09631678b13b05d9d7a-Abstract.html}

\bibitem[{Qi et~al.(2022)Qi, Gao, Hu, Wang, Liu, Bai, Belongie, Yuille, Torr, and Bai}]{qi_occluded_2022}
Qi J, Gao Y, Hu Y, et~al. (2022) Occluded {Video} {Instance} {Segmentation}: {A} {Benchmark}. International Journal of Computer Vision 130(8):2022--2039. \doi{10.1007/s11263-022-01629-1}, \urlprefix\url{https://doi.org/10.1007/s11263-022-01629-1}

\bibitem[{Sahin and Itti(2023)}]{sahin_hoot_2023}
Sahin G, Itti L (2023) {HOOT}: {Heavy} {Occlusions} in {Object} {Tracking} {Benchmark}. In: 2023 {IEEE}/{CVF} {Winter} {Conference} on {Applications} of {Computer} {Vision} ({WACV}). IEEE, Waikoloa, HI, USA, pp 4819--4828, \doi{10.1109/WACV56688.2023.00481}, \urlprefix\url{https://ieeexplore.ieee.org/document/10030507/}

\bibitem[{Simonyan and Zisserman(2015)}]{simonyan_very_2015}
Simonyan K, Zisserman A (2015) Very {Deep} {Convolutional} {Networks} for {Large}-{Scale} {Image} {Recognition}. \urlprefix\url{http://arxiv.org/abs/1409.1556}, arXiv:1409.1556 [cs]

\bibitem[{Tang et~al.(2018)Tang, Schrimpf, Lotter, Moerman, Paredes, Ortega~Caro, Hardesty, Cox, and Kreiman}]{tang_recurrent_2018}
Tang H, Schrimpf M, Lotter W, et~al. (2018) Recurrent computations for visual pattern completion. Proceedings of the National Academy of Sciences 115(35):8835--8840. \doi{10.1073/pnas.1719397115}, \urlprefix\url{https://www.pnas.org/doi/10.1073/pnas.1719397115}, publisher: Proceedings of the National Academy of Sciences

\bibitem[{Touvron et~al.(2021)Touvron, Cord, Douze, Massa, Sablayrolles, and Jegou}]{touvron_training_2021}
Touvron H, Cord M, Douze M, et~al. (2021) Training data-efficient image transformers \& distillation through attention. In: Proceedings of the 38th {International} {Conference} on {Machine} {Learning}. PMLR, pp 10347--10357, \urlprefix\url{https://proceedings.mlr.press/v139/touvron21a.html}, iSSN: 2640-3498

\bibitem[{Tukey(1977)}]{tukey_exploratory_1977}
Tukey JW (1977) Exploratory data analysis. Addison-{Wesley} series in behavioral sciences, Addison-Wesley Pub. Co., Reading, Mass., \urlprefix\url{http://www.gbv.de/dms/bowker/toc/9780201076165.pdf}, oCLC: 3058187

\bibitem[{Wang et~al.(2017)Wang, Xie, Zhang, Zhu, Xie, and Yuille}]{wang_detecting_2017}
Wang J, Xie C, Zhang Z, et~al. (2017) Detecting {Semantic} {Parts} on {Partially} {Occluded} {Objects}. \urlprefix\url{http://arxiv.org/abs/1707.07819}, arXiv:1707.07819 [cs]

\bibitem[{Xiao et~al.(2019)Xiao, Kortylewski, Wu, Qiao, Shen, and Yuille}]{xiao_tdapnet_2019}
Xiao M, Kortylewski A, Wu R, et~al. (2019) {TDAPNet}: {Prototype} {Network} with {Recurrent} {Top}-{Down} {Attention} for {Robust} {Object} {Classification} under {Partial} {Occlusion}. \doi{10.48550/arXiv.1909.03879}, \urlprefix\url{http://arxiv.org/abs/1909.03879}, arXiv:1909.03879 [cs]

\bibitem[{Xiao et~al.(2022)Xiao, Kortylewski, Wu, Qiao, Shen, and Yuille}]{xiao_tdmpnet_2022}
Xiao M, Kortylewski A, Wu R, et~al. (2022) {TDMPNet}: {Prototype} {Network} with {Recurrent} {Top}-{Down} {Modulation} for {Robust} {Object} {Classification} under {Partial} {Occlusion}. \urlprefix\url{https://openreview.net/forum?id=v_KSmk9B5kt}

\bibitem[{Xie et~al.(2017)Xie, Girshick, Dollar, Tu, and He}]{xie_aggregated_2017}
Xie S, Girshick R, Dollar P, et~al. (2017) Aggregated {Residual} {Transformations} for {Deep} {Neural} {Networks}. In: 2017 {IEEE} {Conference} on {Computer} {Vision} and {Pattern} {Recognition} ({CVPR}). IEEE, Honolulu, HI, pp 5987--5995, \doi{10.1109/CVPR.2017.634}, \urlprefix\url{http://ieeexplore.ieee.org/document/8100117/}

\bibitem[{Yang et~al.(2018)Yang, Zhang, Yin, and Liu}]{yang_robust_2018}
Yang HM, Zhang XY, Yin F, et~al. (2018) Robust {Classification} with {Convolutional} {Prototype} {Learning}. In: 2018 {IEEE}/{CVF} {Conference} on {Computer} {Vision} and {Pattern} {Recognition}. IEEE, Salt Lake City, UT, USA, pp 3474--3482, \doi{10.1109/CVPR.2018.00366}, \urlprefix\url{https://ieeexplore.ieee.org/document/8578464/}

\bibitem[{Yun et~al.(2019)Yun, Han, Chun, Oh, Yoo, and Choe}]{yun_cutmix_2019}
Yun S, Han D, Chun S, et~al. (2019) {CutMix}: {Regularization} {Strategy} to {Train} {Strong} {Classifiers} {With} {Localizable} {Features}. In: 2019 {IEEE}/{CVF} {International} {Conference} on {Computer} {Vision} ({ICCV}). IEEE, Seoul, Korea (South), pp 6022--6031, \doi{10.1109/ICCV.2019.00612}, \urlprefix\url{https://ieeexplore.ieee.org/document/9008296/}

\bibitem[{Zhang et~al.(2018)Zhang, Cisse, Dauphin, and Lopez-Paz}]{zhang_mixup_2018}
Zhang H, Cisse M, Dauphin YN, et~al. (2018) mixup: {Beyond} {Empirical} {Risk} {Minimization}. \urlprefix\url{https://openreview.net/forum?id=r1Ddp1-Rb}

\bibitem[{Zhu et~al.(2019)Zhu, Tang, Park, Park, and Yuille}]{zhu_robustness_2019}
Zhu H, Tang P, Park J, et~al. (2019) Robustness of {Object} {Recognition} under {Extreme} {Occlusion} in {Humans} and {Computational} {Models}. \doi{10.48550/arXiv.1905.04598}, \urlprefix\url{http://arxiv.org/abs/1905.04598}, arXiv:1905.04598 [cs]

\end{thebibliography}

\end{document}